%% file: main.tex
\documentclass{article}
\usepackage{ijcai25}

\usepackage[table,x11names,dvipsnames]{xcolor}         % colors

\usepackage{times}
\usepackage{soul}
\usepackage{url}
\usepackage[hidelinks]{hyperref}
\usepackage[utf8]{inputenc}
\usepackage[small]{caption}
\usepackage{graphicx}
\usepackage{amsmath}
\usepackage{amsthm}
\usepackage{booktabs}
\usepackage{algorithm}
\usepackage{algorithmic}
\usepackage[switch]{lineno}

\title{
Advancing Generalization Across a Variety of Abstract Visual Reasoning Tasks
}

\author{
Mikołaj Małkiński$^1$
\And
Jacek Mańdziuk$^{1,2}$\\
\affiliations
$^1$Warsaw University of Technology, Warsaw, Poland\\
$^2$AGH University of Krakow, Krakow, Poland\\
\emails
mikolaj.malkinski.dokt@pw.edu.pl
$\diamond$
jacek.mandziuk@pw.edu.pl
}
\input{sections/preamble}

\begin{document}

% Keywords: Abstract Visual Reasoning, Raven's Progressive Matrices, Deep Learning, Neural Networks

\maketitle

\begin{abstract}
\input{sections/abstract}
\end{abstract}

\input{sections/introduction}
\input{sections/related-work}

\input{sections/method}
\input{sections/experiments}
\input{sections/conclusion}

\appendix

% \section*{Ethical Statement}

% There are no ethical issues.

\input{sections/acknowledgment}

\bibliographystyle{named}
\bibliography{main}

\newpage
\input{sections/appendix}

\end{document}

%% file: sections/preamble.tex
\usepackage{amsmath}  % for the "cases" environment
\usepackage{amssymb}
\usepackage{bm}
\usepackage{bbm}
\usepackage{multirow}  % span table cell from multiple rows
\usepackage{makecell}  % advanced table cell formatting
\usepackage{textcomp}  % cross symbol in table
\usepackage{numprint}  % nicely format big numbers
\usepackage{caption}
\usepackage{subcaption}
\usepackage{float}  % force table placement with [H]
\usepackage[above]{placeins}  % for \FloatBarrier
\usepackage{soul}  % for strike-through: \st{}
\npthousandsep{,}
\usepackage{wrapfig}   % wrap text around a figure or table
\usepackage{tikz}
\usetikzlibrary{calc,shapes.geometric,trees,positioning,arrows,arrows.meta,backgrounds,fit,decorations.pathreplacing,decorations.markings,calligraphy,matrix}
\usepackage{booktabs}       % professional-quality tables

\usepackage{setspace}
\usepackage[colaction]{multicol}

\makeatletter
\DeclareRobustCommand{\rvdots}{%
  \vbox{
    \baselineskip2.5\p@\lineskiplimit\z@
    \kern-\p@
    \hbox{.}\hbox{.}\hbox{.}
  }%
}
\makeatother

\definecolor{spectral-1}{RGB}{158, 1, 66}
\definecolor{spectral-2}{RGB}{177, 22, 70}
\definecolor{spectral-3}{RGB}{198, 46, 75}
\definecolor{spectral-4}{RGB}{217, 68, 77}
\definecolor{spectral-5}{RGB}{229, 86, 72}
\definecolor{spectral-6}{RGB}{241, 105, 67}
\definecolor{spectral-7}{RGB}{246, 129, 76}
\definecolor{spectral-8}{RGB}{250, 152, 86}
\definecolor{spectral-9}{RGB}{253, 176, 99}
\definecolor{spectral-10}{RGB}{253, 196, 115}
\definecolor{spectral-11}{RGB}{253, 216, 132}
\definecolor{spectral-12}{RGB}{254, 231, 151}
\definecolor{spectral-13}{RGB}{254, 243, 171}
\definecolor{spectral-14}{RGB}{254, 254, 190}
\definecolor{spectral-15}{RGB}{245, 251, 176}
\definecolor{spectral-16}{RGB}{235, 247, 161}
\definecolor{spectral-17}{RGB}{220, 241, 153}
\definecolor{spectral-18}{RGB}{197, 231, 158}
\definecolor{spectral-19}{RGB}{174, 222, 163}
\definecolor{spectral-20}{RGB}{148, 212, 164}
\definecolor{spectral-21}{RGB}{123, 202, 164}
\definecolor{spectral-22}{RGB}{97, 189, 166}
\definecolor{spectral-23}{RGB}{77, 166, 176}
\definecolor{spectral-24}{RGB}{57, 143, 185}
\definecolor{spectral-25}{RGB}{61, 121, 182}
\definecolor{spectral-26}{RGB}{78, 99, 171}
\definecolor{spectral-27}{RGB}{94, 79, 162}

\definecolor{spectral-1-1}{RGB}{158, 1, 66}
\definecolor{spectral-1-2}{RGB}{164, 8, 67}
\definecolor{spectral-1-3}{RGB}{170, 15, 69}
\definecolor{spectral-1-4}{RGB}{177, 22, 70}
\definecolor{spectral-1-5}{RGB}{183, 29, 72}
\definecolor{spectral-1-6}{RGB}{192, 39, 74}
\definecolor{spectral-1-7}{RGB}{198, 46, 75}
\definecolor{spectral-1-8}{RGB}{205, 53, 77}
\definecolor{spectral-1-9}{RGB}{211, 60, 78}
\definecolor{spectral-1-10}{RGB}{216, 66, 77}
\definecolor{spectral-1-11}{RGB}{220, 73, 75}
\definecolor{spectral-1-12}{RGB}{224, 79, 74}
\definecolor{spectral-1-13}{RGB}{228, 85, 73}
\definecolor{spectral-1-14}{RGB}{231, 90, 71}
\definecolor{spectral-1-15}{RGB}{235, 96, 70}
\definecolor{spectral-1-16}{RGB}{240, 103, 68}
\definecolor{spectral-1-17}{RGB}{244, 109, 67}
\definecolor{spectral-1-18}{RGB}{245, 116, 70}
\definecolor{spectral-1-19}{RGB}{246, 124, 74}
\definecolor{spectral-1-20}{RGB}{247, 131, 77}
\definecolor{spectral-1-21}{RGB}{248, 142, 82}
\definecolor{spectral-1-22}{RGB}{249, 149, 85}
\definecolor{spectral-1-23}{RGB}{250, 157, 89}
\definecolor{spectral-1-24}{RGB}{251, 165, 92}
\definecolor{spectral-1-25}{RGB}{252, 172, 96}
\definecolor{spectral-1-26}{RGB}{253, 180, 102}
\definecolor{spectral-1-27}{RGB}{253, 186, 107}

\definecolor{spectral-2-1}{RGB}{253, 192, 112}
\definecolor{spectral-2-2}{RGB}{253, 198, 117}
\definecolor{spectral-2-3}{RGB}{253, 204, 122}
\definecolor{spectral-2-4}{RGB}{253, 212, 129}
\definecolor{spectral-2-5}{RGB}{253, 218, 134}
\definecolor{spectral-2-6}{RGB}{254, 224, 139}
\definecolor{spectral-2-7}{RGB}{254, 227, 145}
\definecolor{spectral-2-8}{RGB}{254, 231, 151}
\definecolor{spectral-2-9}{RGB}{254, 236, 159}
\definecolor{spectral-2-10}{RGB}{254, 239, 165}
\definecolor{spectral-2-11}{RGB}{254, 243, 171}
\definecolor{spectral-2-12}{RGB}{254, 247, 177}
\definecolor{spectral-2-13}{RGB}{254, 250, 183}
\definecolor{spectral-2-14}{RGB}{254, 254, 190}
\definecolor{spectral-2-15}{RGB}{251, 253, 185}
\definecolor{spectral-2-16}{RGB}{248, 252, 181}
\definecolor{spectral-2-17}{RGB}{245, 251, 176}
\definecolor{spectral-2-18}{RGB}{242, 250, 171}
\definecolor{spectral-2-19}{RGB}{238, 248, 165}
\definecolor{spectral-2-20}{RGB}{235, 247, 161}
\definecolor{spectral-2-21}{RGB}{232, 246, 156}
\definecolor{spectral-2-22}{RGB}{230, 245, 152}
\definecolor{spectral-2-23}{RGB}{223, 242, 153}
\definecolor{spectral-2-24}{RGB}{213, 238, 155}
\definecolor{spectral-2-25}{RGB}{206, 235, 156}
\definecolor{spectral-2-26}{RGB}{199, 232, 158}
\definecolor{spectral-2-27}{RGB}{192, 229, 159}

\definecolor{spectral-3-1}{RGB}{186, 227, 160}
\definecolor{spectral-3-2}{RGB}{176, 223, 162}
\definecolor{spectral-3-3}{RGB}{169, 220, 164}
\definecolor{spectral-3-4}{RGB}{161, 217, 164}
\definecolor{spectral-3-5}{RGB}{153, 214, 164}
\definecolor{spectral-3-6}{RGB}{145, 210, 164}
\definecolor{spectral-3-7}{RGB}{134, 206, 164}
\definecolor{spectral-3-8}{RGB}{126, 203, 164}
\definecolor{spectral-3-9}{RGB}{118, 200, 164}
\definecolor{spectral-3-10}{RGB}{110, 197, 164}
\definecolor{spectral-3-11}{RGB}{102, 194, 165}
\definecolor{spectral-3-12}{RGB}{93, 184, 168}
\definecolor{spectral-3-13}{RGB}{87, 178, 171}
\definecolor{spectral-3-14}{RGB}{81, 171, 174}
\definecolor{spectral-3-15}{RGB}{75, 164, 177}
\definecolor{spectral-3-16}{RGB}{69, 157, 180}
\definecolor{spectral-3-17}{RGB}{61, 148, 183}
\definecolor{spectral-3-18}{RGB}{55, 141, 186}
\definecolor{spectral-3-19}{RGB}{50, 134, 188}
\definecolor{spectral-3-20}{RGB}{56, 128, 185}
\definecolor{spectral-3-21}{RGB}{61, 121, 182}
\definecolor{spectral-3-22}{RGB}{68, 112, 177}
\definecolor{spectral-3-23}{RGB}{73, 105, 174}
\definecolor{spectral-3-24}{RGB}{78, 99, 171}
\definecolor{spectral-3-25}{RGB}{83, 92, 168}
\definecolor{spectral-3-26}{RGB}{88, 85, 165}
\definecolor{spectral-3-27}{RGB}{94, 79, 162}

%% file: sections/abstract.tex
The abstract visual reasoning (AVR) domain presents a diverse suite of analogy-based tasks devoted to studying model generalization.
Recent years have brought dynamic progress in the field, particularly in i.i.d. scenarios, in which models are trained and evaluated on the same data distributions.
Nevertheless, o.o.d. setups that assess model generalization to new test distributions remain challenging even for the most recent models.
To advance generalization in AVR tasks, we present the \emph{Pathways of Normalized Group Convolution} model (PoNG), a novel neural architecture that features group convolution, normalization, and a parallel design.
We consider a wide set of AVR benchmarks, including Raven's Progressive Matrices and visual analogy problems with both synthetic and real-world images.
The experiments demonstrate strong generalization capabilities of the proposed model, which in several settings outperforms the existing literature methods.

%% file: sections/introduction.tex
\section{Introduction}
\label{sec:introduction}

The abstract visual reasoning (AVR) domain encompasses visual tasks requiring reasoning about abstract patterns expressed through image-based analogies.
A classical AVR task, Raven's Progressive Matrices (RPMs)~\cite{raven1936mental,raven1998raven}, illustrated in Fig.~\ref{fig:rpms}, consists of a $3\times3$ grid of panels with the bottom-right panel missing.
Panels in the first two rows are designed according to some number of abstract rules that govern objects and attributes in the images.
The task is to complete the grid by selecting the correct answer from the eight provided choices.
Another AVR task, visual analogies, shown in Fig.~\ref{fig:vaps}, involves two rows of images.
The top row presents an abstract relation that must be instantiated in the bottom row by selecting one of the four answer panels that correctly completes the analogy.

\input{figures/overview}

Solving AVR tasks involves detecting rule patterns across images, abstracting them into crisp concepts, and applying these concepts to novel scenarios.
For example, matrices in the visual analogy problems (VAP) dataset~\cite{hill2018learning} present different domains in matrix rows (e.g., \texttt{shape type} (top) and \texttt{line type} (bottom) in Fig.~\ref{fig:vaps-vap}), emphasizing the importance of forming domain-independent concept representations.
Such analogy-making abilities are closely tied to fluid intelligence~\cite{snow1984topography,carpenter1990one,lake2017building}, a cornerstone of human cognition.
Replicating these capabilities in learning systems has been a long-standing goal of research in the field~\cite{gentner1980structure,hofstadter1995fluid,french2002computational,lovett2007analogy,gentner2011computational}.

A key aspect of AVR tasks, central to our work, are their systematic problem generation methods.
Underneath each AVR task design lies a precise definition of its abstract structure, which defines the rule patterns expressed in the matrices.
For instance, each PGM matrix~\cite{santoro2018measuring}
has a corresponding abstract structure $\mathcal{S} = \{(r, o, a)\ \vert\ r\in\mathcal{R}, o\in\mathcal{O}, a\in\mathcal{A}\}$, where $\mathcal{R}$ $=$ $\{\texttt{progression},$ $\texttt{XOR},$ $\texttt{OR},$ $\texttt{AND},$ $\texttt{consistent union}\}$, $\mathcal{O}$ $=$ $\{\texttt{shape},$ $\texttt{line}\}$, and $\mathcal{A}$ $=$ $\{\texttt{size},$ $\texttt{type},$ $\texttt{color},$ $\texttt{position},$ $\texttt{number}\}$ are the sets of rules, objects, and attributes, resp.
% The abstract structure of the PGM matrix in Fig.~\ref{fig:rpm-pgm} is $\mathcal{S}$ $=$ $\{(\texttt{line},$ $\texttt{type},$ $\texttt{consistent union}),$ $(\texttt{shape},$ $\texttt{number},$ $\texttt{progression})\}$, with the first triple governing matrix columns and the second one concerning matrix rows.
This formal specification facilitates defining dataset splits with varying feature distributions, enabling the evaluation of generalization by training models on matrices with specific abstract structures and testing them on matrices with different structures. 

\input{figures/a-i-raven}
\input{figures/i-raven-mesh}

\paragraph{Motivation.}
Several AVR studies have addressed the i.i.d. problem formulation, where models are trained and tested on matrices sampled from a shared feature distribution.
Continuous improvements have produced methods that surpass human performance on tasks like RPMs, given sufficient amount of training data~\cite{hernandez2016computer,malkinski2022deep}.
Other research lines have demonstrated the benefits of knowledge transfer~\cite{mandziuk2019deepiq,tomaszewska2022duel} and multi-task learning~\cite{malkinski2024one}.
Despite these achievements, o.o.d. problem formulations, where models are evaluated on matrices sampled from a different feature distribution than the one used for training, remain a major challenge even for state-of-the-art (SOTA) deep learning (DL) models.
Moreover, existing approaches in the AVR domain primarily target synthetic tasks with simple 2D geometric shapes, without considering their applicability to problems with real-world data.
In this work, we strive to develop a model architecture that not only performs well in i.i.d. tasks but also excels in o.o.d. settings.
Additionally, we consider both synthetic and real-world setups to broaden the applicability of the proposed approach.

\paragraph{Contribution.}
To tackle these open challenges, we introduce the following contributions:
\begin{itemize}
    \item We propose \emph{Pathways of Normalized Group Convolution} (PoNG), a new neural model for AVR tasks that integrates group convolution, normalization, and a parallel design.
    \item We perform a comprehensive evaluation of PoNG against a wide range of SOTA models. The experiments show PoNG's versatility in both i.i.d. and o.o.d. problem setups, spanning RPMs and VAPs in synthetic and real-world scenarios.
    \item We conduct an ablation study to analyze the contributions of PoNG's modules, providing deeper insights into its design.
\end{itemize}

%% file: figures/overview.tex
\begin{figure}[t]
  \centering
  \begin{subfigure}{0.224\textwidth}
    \centering
    \includegraphics[width=\textwidth]{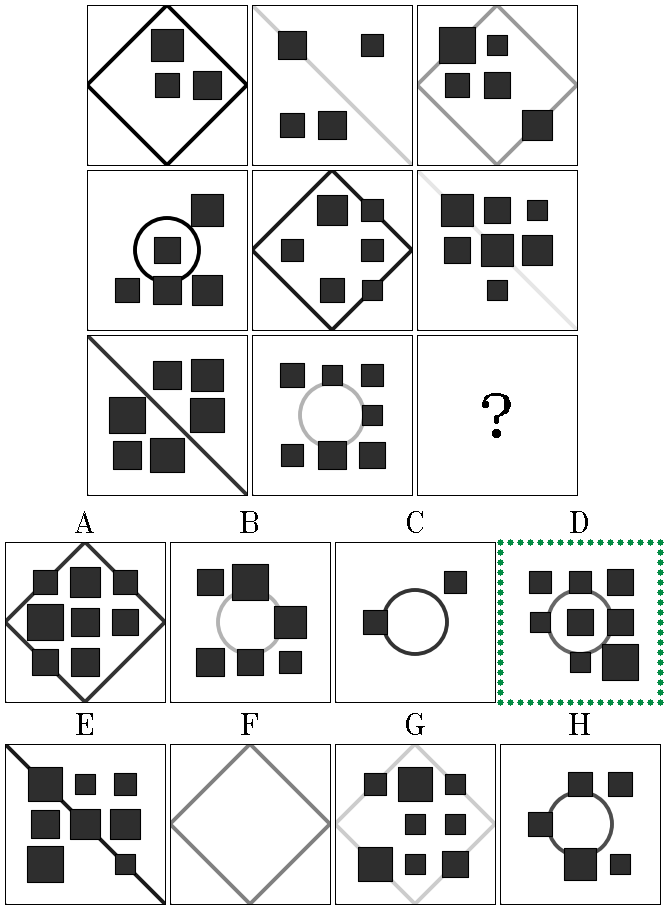}
    \caption{PGM~\protect\cite{santoro2018measuring}}
    \label{fig:rpm-pgm}
  \end{subfigure}
  \hfill
  \begin{subfigure}{0.224\textwidth}
    \centering
    \includegraphics[width=\textwidth]{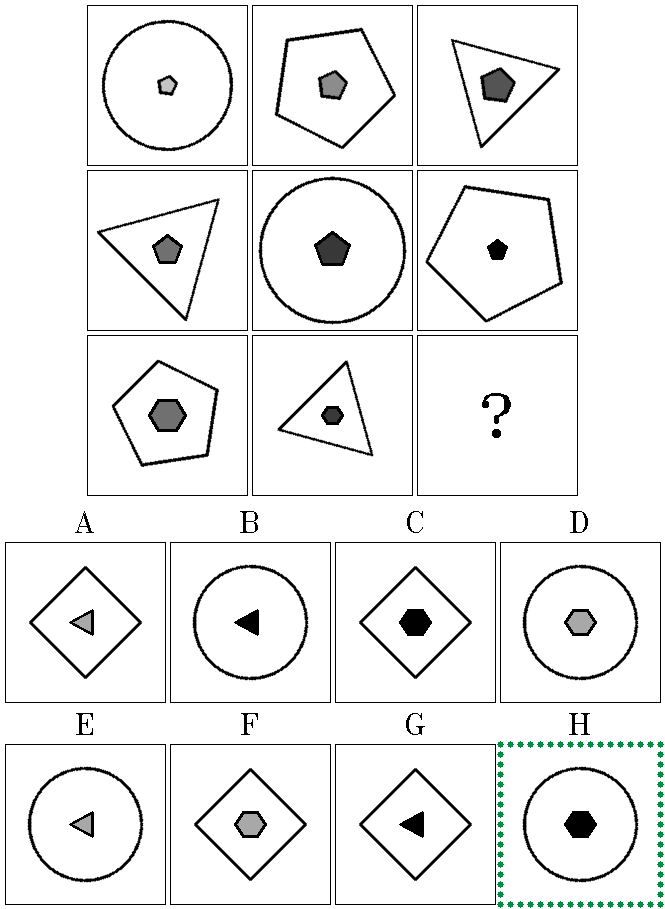}
    \caption{I-RAVEN~\protect\cite{hu2021stratified}}
    \label{fig:rpm-iraven}
  \end{subfigure}
  \caption{Raven's Progressive Matrices (RPMs).}
  \label{fig:rpms}
\end{figure}
\begin{figure}[t]
  \centering
  \begin{subfigure}[b]{0.22\textwidth}
    \centering
    \includegraphics[width=\textwidth]{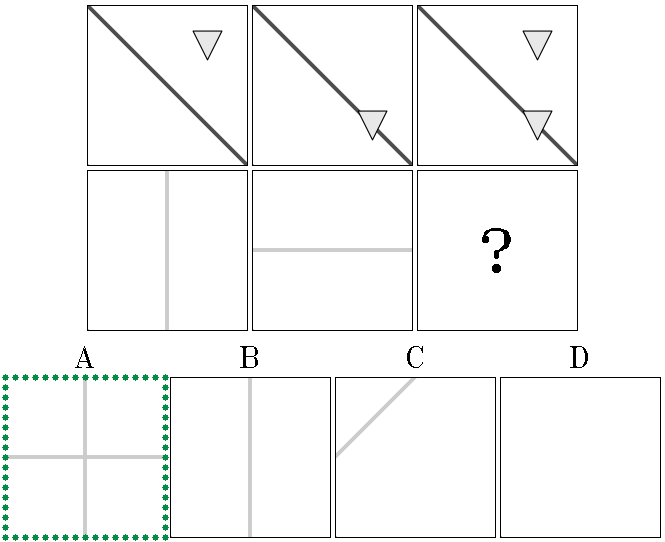}
    \caption{VAP~\protect\cite{hill2018learning}}
    \label{fig:vaps-vap}
  \end{subfigure}
  \hfil
  \begin{subfigure}[b]{0.255\textwidth}
    \centering
    \includegraphics[width=\textwidth]{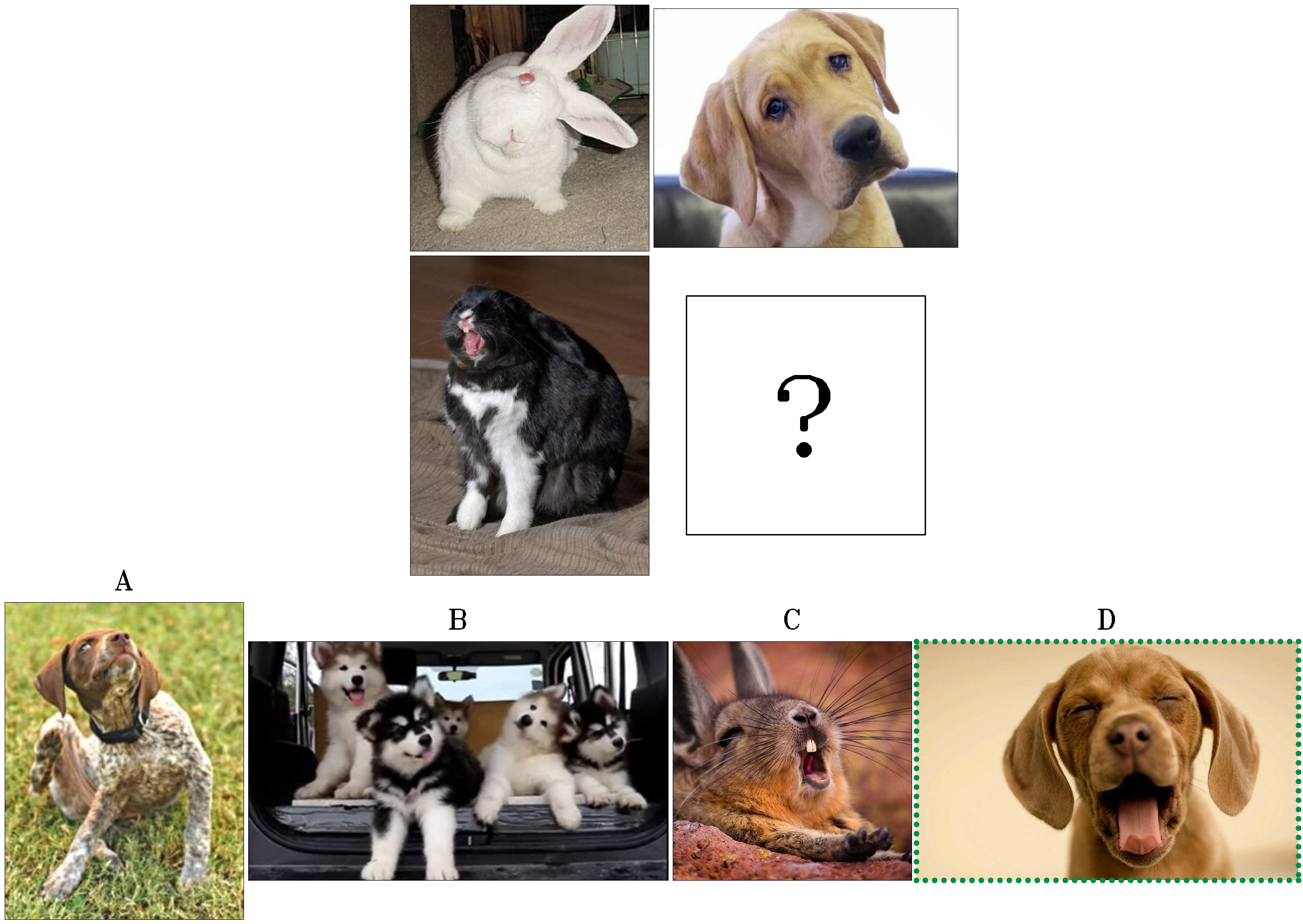}
    \caption{VASR~\protect\cite{bitton2023vasr}}
    \label{fig:vaps-vasr}
  \end{subfigure}
  \caption{Visual analogies.}
  \label{fig:vaps}
\end{figure}

%% file: figures/a-i-raven.tex
\begin{figure}[t]
  \centering
  \begin{subfigure}{0.224\textwidth}
    \centering
    \includegraphics[width=\textwidth]{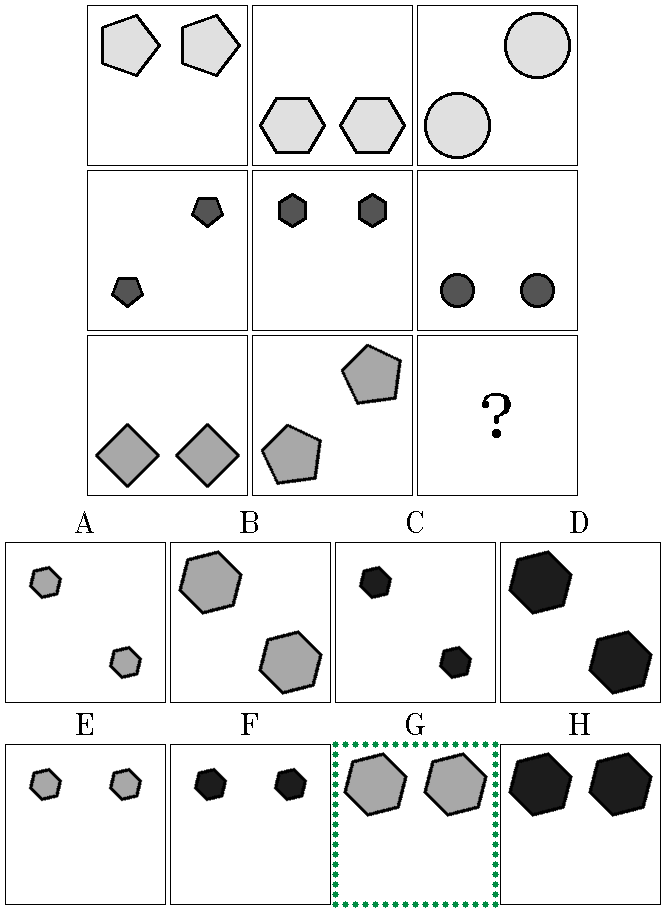}
    \caption{\texttt{A/ColorSize} train}
    \label{fig:a-i-raven-train}
  \end{subfigure}
  \hfil
  \begin{subfigure}{0.224\textwidth}
    \centering
    \includegraphics[width=\textwidth]{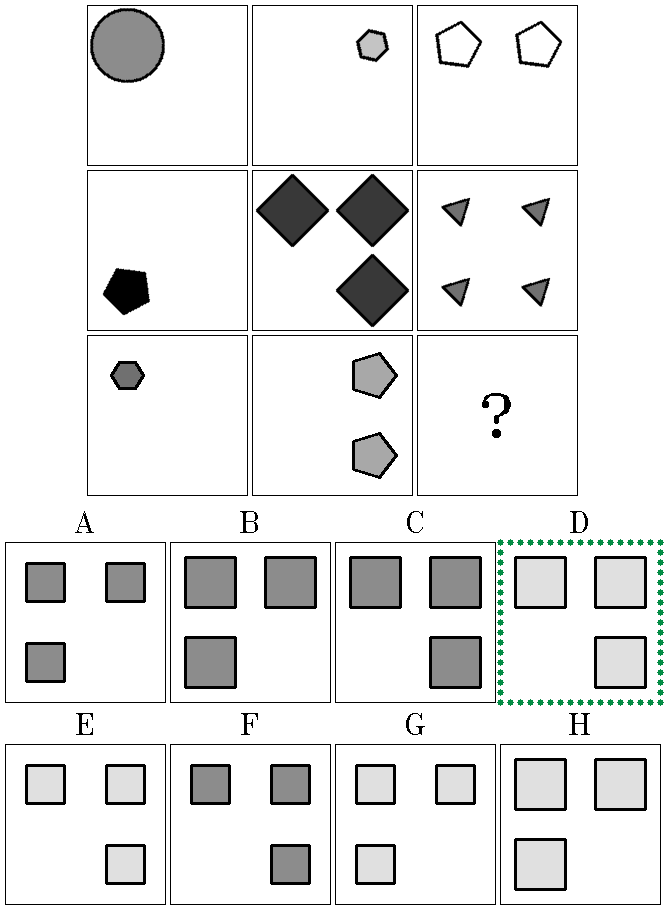}
    \caption{\texttt{A/ColorSize} test}
    \label{fig:a-i-raven-test}
  \end{subfigure}
  \caption{A-I-RAVEN~\protect\cite{malkinski2025airaven}.}
  \label{fig:a-i-raven}
\end{figure}

%% file: figures/i-raven-mesh.tex
\begin{figure}[t]
  \centering
  \begin{subfigure}{0.224\textwidth}
    \centering
    \includegraphics[width=\textwidth]{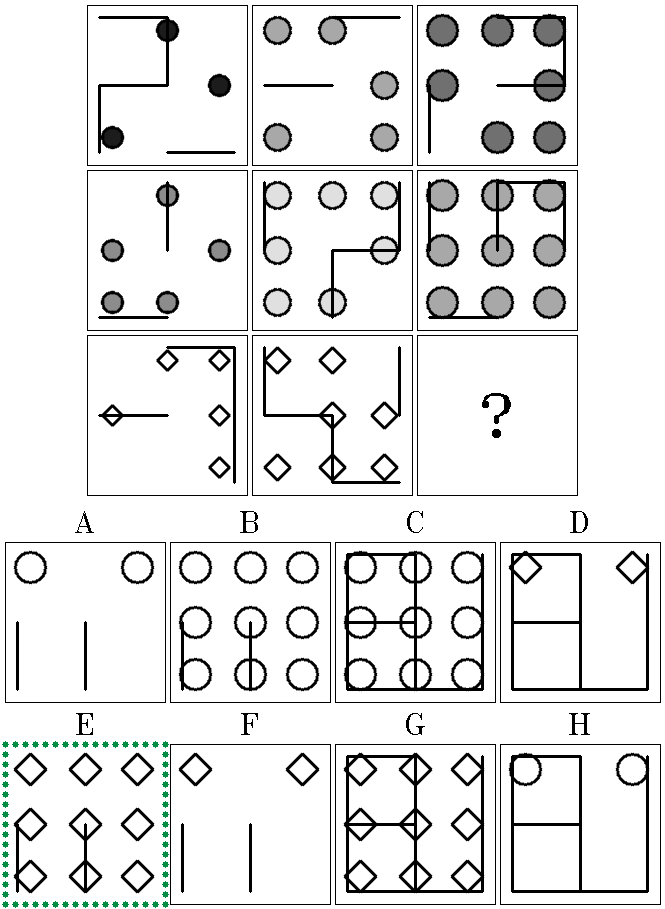}
    \caption{Number DistributeThree}
    \label{fig:i-raven-mesh-number}
  \end{subfigure}
  \hfil
  \begin{subfigure}{0.224\textwidth}
    \centering
    \includegraphics[width=\textwidth]{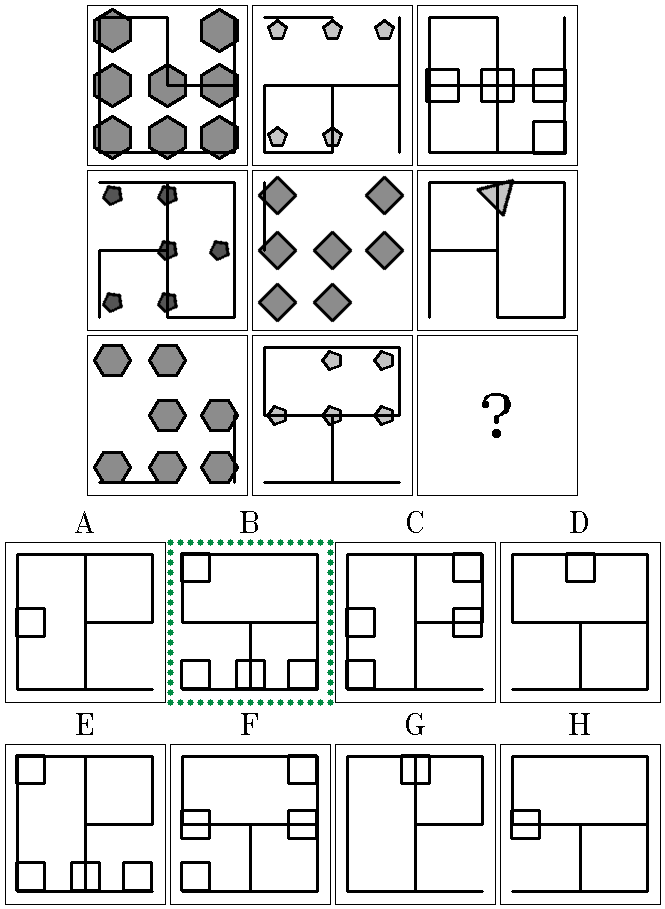}
    \caption{Position Arithmetic}
    \label{fig:i-raven-mesh-position}
  \end{subfigure}
  \caption{I-RAVEN-Mesh~\protect\cite{malkinski2025airaven}.}
  \label{fig:i-raven-mesh}
\end{figure}

%% file: sections/related-work.tex
\section{Related Work}
\label{sec:related-work}

\paragraph{AVR tasks.}
The AVR domain comprises a wide range of challenges~\cite{mitchell2021abstraction,van2021much,stabinger2021evaluating,malkinski2022review}.
Most relevant to our work are tasks involving RPMs and visual analogies.
After the introduction of early RPM datasets~\cite{matzen2010recreating,wang2015automatic,hoshen2017iq}, two large-scale benchmarks were developed and broadly adopted in the DL literature.
PGM~\cite{santoro2018measuring} (Fig.~\ref{fig:rpm-pgm}) introduced $8$ regimes to measure generalization of DL models.
RAVEN~\cite{zhang2019raven} presented matrices with hierarchical structures across $7$ figure configurations.
Subsequent works further expanded the RAVEN dataset line:
I-RAVEN~\cite{hu2021stratified} (Fig.~\ref{fig:rpm-iraven}) mitigated a bias in RAVEN's answer generation method, A-I-RAVEN~\cite{malkinski2025airaven} (Fig.~\ref{fig:a-i-raven}) defined $10$ generalization regimes of varying complexity, and I-RAVEN-Mesh~\cite{malkinski2025airaven} (Fig.~\ref{fig:i-raven-mesh}) overlayed line-based patterns on top of the matrices.
A parallel research stream introduced the VAP dataset~\cite{hill2018learning} (Fig.~\ref{fig:vaps-vap}), which similarly to PGM enables measuring generalization on matrices with a different structure.
VASR~\cite{bitton2023vasr} (Fig.~\ref{fig:vaps-vasr}) introduced analogies with real-world images requiring understanding of rich visual scenes.
A detailed description of the datasets used in this work is provided in Section~\ref{sec:experiments-tasks}.

\input{figures/model}

\paragraph{AVR solvers.}
Early attempts to solve AVR tasks with DL models were prompted by the development of Relation Network (RN)~\cite{santoro2017simple}.
WReN~\cite{santoro2018measuring} applies RN to panel embeddings, CoPINet~\cite{zhang2019learning} integrates RN with contrastive mechanisms, and MRNet~\cite{benny2020scale} embeds RN into a multi-scale architecture.
Differently, SRAN~\cite{hu2021stratified} processes distinct panel groups with dedicated ResNet encoders, SCL~\cite{wu2020scattering} splits embeddings into groups processed by a shared neural layer, RelBase~\cite{spratley2020closer} primarily relies on convolutional layers, ARII~\cite{zhang2022learningrobust} learns robust rule representations via internal inferences, CPCNet~\cite{yang2023cognitively} utilizes a self-contrasting learning process to align perceptual and conceptual input representations, PredRNet~\cite{yang2023neural} mimics the prediction and matching process, and DRNet~\cite{zhao2024learning} merges panel representations of two independent encoders.
Other works develop neuro-symbolic approaches~\cite{zhang2021abstract,zhang2022learning} or perform explicit object recognition prior to reasoning~\cite{mondal2023learning,mondal2024slot}.
Neural Structure Mapping~\cite{shekhar2022neural} decouples perception from reasoning to solve visual analogies.
\citeauthor{bitton2023vasr}~[\citeyear{bitton2023vasr}] formulate several zero-shot and supervised methods to solve real-world analogies using a frozen pre-trained Vision Transformer~\cite{dosovitskiy2021an} as the panel encoder.
Although diverse approaches have been tried to tackle AVR benchmarks, contemporary models continue to exhibit limitations in generalization.
In this context, we propose PoNG, a new versatile AVR model that performs well across diverse tasks.

%% file: figures/model.tex
\newcommand{\unit}{40pt}
\newcommand{\rec}[4][]{
    \pgfmathsetmacro \a {0.12} % side lengths
    \pgfmathsetmacro \x {{#3*\a-\a}}
    \pgfmathsetmacro \y {{-#2*\a+\a}}

    \draw[fill=#4, very thin] (\x,\y) -- (\x+\a,\y) -- (\x+\a,\y-\a) -- (\x,\y-\a) -- cycle;
    \node[inner sep=0, anchor=center] at (\x+\a*0.5, \y-\a*0.5) {\tiny#1};
}

\newcommand{\rpm}[1]{\includegraphics[width=0.04\textwidth]{images/rpm/context_#1}}

\definecolor{layer}{HTML}{E1E1E1}
\definecolor{activation}{HTML}{E69F00}
\definecolor{conv}{HTML}{56B4E9}
\definecolor{linear}{HTML}{009E73}
\definecolor{norm}{HTML}{CC79A7}
\definecolor{pathways}{HTML}{D55E00}

\tikzstyle{activation} = [fill=activation!20]
\tikzstyle{conv} = [fill=conv!20]
\tikzstyle{norm} = [fill=norm!20]
\tikzstyle{linear} = [fill=linear!20]
\tikzstyle{pathways} = [fill=pathways!20]
\tikzstyle{bottleneck} = [fill=bottleneck!20]

\tikzstyle{layer} = [rectangle, rounded corners=2pt, text centered, draw=black, fill=layer, inner sep=0, minimum height=0.2*\unit, minimum width=1*\unit, align=center, font={\scriptsize}]
\tikzstyle{rotated-layer} = [rectangle, rounded corners=2pt, text centered, draw=black, fill=layer, inner sep=0, minimum height=1*\unit, minimum width=0.2*\unit, align=center, font={\scriptsize}]
\tikzstyle{empty-layer} = [rectangle, inner sep=0, minimum height=0.2*\unit, minimum width=1*\unit]
\tikzstyle{empty-rotated-layer} = [rectangle, inner sep=0, minimum height=1*\unit, minimum width=0.2*\unit]
\tikzstyle{label} = [align=center, font={\scriptsize}]
\tikzstyle{rotated-label} = [align=center, font={\scriptsize}, rotate=270]
\tikzstyle{text-label} = [align=center, font={\scriptsize}, inner sep=1pt]
\tikzstyle{caption} = [align=center, font={\footnotesize}, inner sep=1pt]
\tikzstyle{rotated-text-label} = [align=center, font={\scriptsize}, rotate=270, inner sep=1pt]
\tikzstyle{resplus} = [circle, draw=black, inner sep=0]
\tikzstyle{arrow} = [-Stealth]
\tikzstyle{curved-line} = [rounded corners=5pt]
\tikzstyle{background} = [draw, fill=black!10, rounded corners=5pt, densely dashed, inner sep=0.05*\unit]
\tikzstyle{panel} = [thick, rectangle, draw=black, inner sep=0]
\tikzstyle{embedding} = [rectangle, rounded corners=1pt, text centered, draw=black, fill=layer!50, inner sep=0, minimum height=0.2*\unit, minimum width=0.2*\unit]
\tikzstyle{rotated-embedding} = [rectangle, rounded corners=1pt, text centered, draw=black, fill=layer!50, inner sep=0, minimum height=0.4*\unit, minimum width=0.2*\unit]
\tikzstyle{empty-rotated-embedding} = [rectangle, rounded corners=1pt, inner sep=0, minimum height=0.4*\unit, minimum width=0.2*\unit]

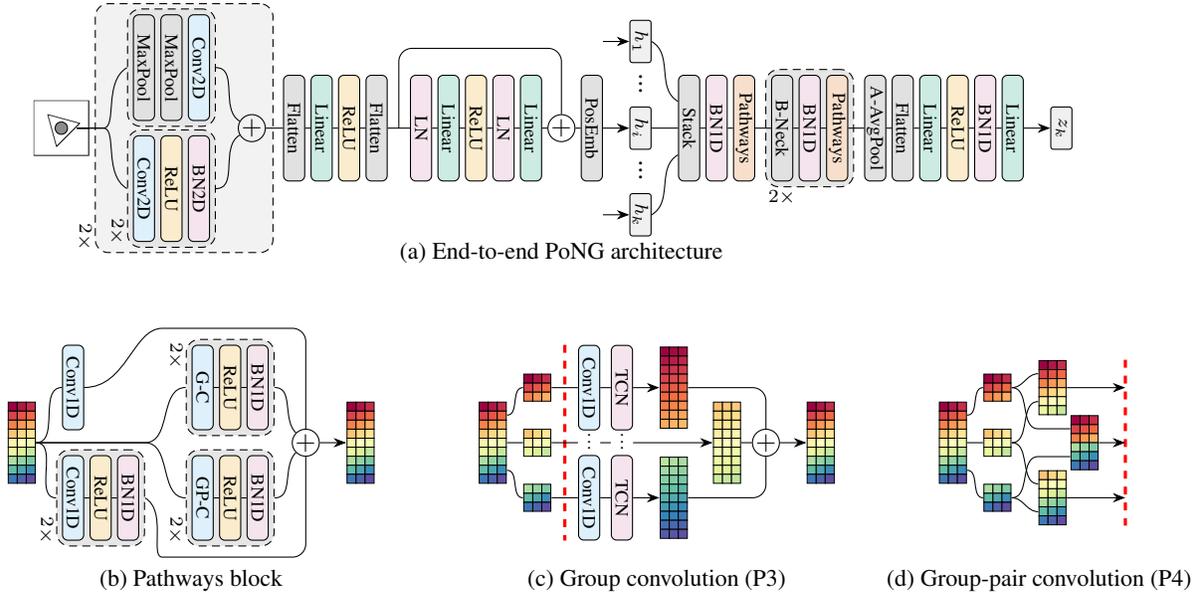
\begin{figure*}[t]
    \centering
    \begin{tikzpicture}[node distance = 2cm, auto]

        % === Model architecture ===

        % Panel encoder
        
        \begin{scope}[node distance = 0.5*\unit]
            \node (panel) [panel] {\rpm{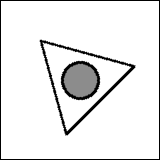}};
            \node (m12-start) [inner sep=0, right=of panel] {};
        \end{scope}

        \begin{scope}[node distance = 0.05*\unit]
            \node (m1-conv) [rotated-layer, conv, below=of m12-start] {};
            \node (m1-relu) [rotated-layer, activation, right=of m1-conv] {};
            \node (m1-bn) [rotated-layer, norm, right=of m1-relu] {};
        \end{scope}
        \node [rotated-label] at (m1-conv) {Conv2D};
        \node [rotated-label] at (m1-relu) {ReLU};
        \node [rotated-label] at (m1-bn) {BN2D};

        \begin{scope}[node distance = 0.05*\unit]
            \node (m2-maxpool1) [rotated-layer, above=of m12-start] {};
            \node (m2-maxpool2) [rotated-layer, right=of m2-maxpool1] {};
            \node (m2-conv) [rotated-layer, conv, right=of m2-maxpool2] {};
        \end{scope}
        \node [rotated-label] at (m2-maxpool1) {MaxPool};
        \node [rotated-label] at (m2-maxpool2) {MaxPool};
        \node [rotated-label] at (m2-conv) {Conv2D};

        \begin{scope}[node distance = 0.35*\unit]
            \node (m12-resplus-start) [inner sep=0] at ($ (m2-conv)!0.5!(m1-bn) $) {};
            \node (m12-resplus) [resplus, right=of m12-resplus-start] {$+$};
        \end{scope}

        % Panel encoder post-processing
        \begin{scope}[node distance = 0.05*\unit]
            \node (m3-flat1) [rotated-layer, xshift=0.4*\unit] at (m12-resplus) {};
            \node (m3-linear1) [rotated-layer, linear, right=of m3-flat1] {};
            \node (m3-relu1) [rotated-layer, activation, right=of m3-linear1] {};
            \node (m3-flat2) [rotated-layer, right=of m3-relu1] {};
            \node (m3-dummy1) [empty-rotated-layer, minimum width=0.1*\unit, right=of m3-flat2] {};
            \node (m3-ln1) [rotated-layer, norm, right=of m3-dummy1] {};
            \node (m3-linear2) [rotated-layer, linear, right=of m3-ln1] {};
            \node (m3-relu2) [rotated-layer, activation, right=of m3-linear2] {};
            \node (m3-ln2) [rotated-layer, norm, right=of m3-relu2] {};
            \node (m3-linear3) [rotated-layer, linear, right=of m3-ln2] {};
            \node (m3-resplus) [resplus, right=of m3-linear3] {$+$};
            \node (m3-position) [rotated-layer, right=of m3-resplus] {};
            \node (m3-h) [rotated-embedding, right=of m3-position, xshift=0.2*\unit] {};
        \end{scope}
        \node [rotated-label] at (m3-flat1) {Flatten};
        \node [rotated-label] at (m3-linear1) {Linear};
        \node [rotated-label] at (m3-relu1) {ReLU};
        \node [rotated-label] at (m3-flat2) {Flatten};
        \node [rotated-label] at (m3-ln1) {LN};
        \node [rotated-label] at (m3-linear2) {Linear};
        \node [rotated-label] at (m3-relu2) {ReLU};
        \node [rotated-label] at (m3-ln2) {LN};
        \node [rotated-label] at (m3-linear3) {Linear};
        \node [rotated-label] at (m3-position) {PosEmb};
        \node [rotated-label] at (m3-h) {$h_i$};

        % Panel encoder backgrounds
        \node (m12-l-start) [inner sep=0] at ($ (m1-conv.north west)!0.5!(m2-maxpool1.south west) $) {};
        \node (m12-l-split) [inner sep=0] at ($ (panel)!2/5!(m12-l-start) $) {};

        \begin{scope}[on background layer]
            \node (m12-bg) [fit={($ (m12-l-split.center) + (0.05*\unit, 0) $) ($ (m1-conv.south) + (0, -0.1*\unit) $) ($ (m2-maxpool1.north) + (0, 0.1*\unit) $) ($ (m12-resplus)!1/2!(m3-flat1) $)}, background, inner sep=0, fill=black!5] {};
        \end{scope}
        \node (m12-2x) [rotated-text-label, anchor=north east] at (m12-bg.south west) {$2 \times$};

        \begin{scope}[on background layer]
            \node (m1-bg) [fit={(m1-conv) (m1-bn)}, background] {};
        \end{scope}
        \node (m1-2x) [rotated-text-label, anchor=north east] at (m1-bg.south west) {$2 \times$};
        
        \begin{scope}[on background layer]
            \node (m2-bg) [fit={(m2-maxpool1) (m2-conv)}, background] {};
        \end{scope}

        % Panel encoder lines
        \node (m12-l-start) [inner sep=0] at ($ (m1-bg.north west)!0.5!(m2-bg.south west) $) {};
        \draw [curved-line] (panel) -- (m12-l-split.center) to[out=0, in=210] (m2-bg.west);
        \draw [curved-line] (panel) -- (m12-l-split.center) to[out=0, in=150] (m1-bg.west);
        \draw (m2-bg) to[out=0, in=150] (m12-resplus);
        \draw (m1-bg) to[out=0, in=210] (m12-resplus);

        \draw (m12-resplus) -- (m3-flat1) -- (m3-linear1) -- (m3-relu1) -- (m3-flat2) -- (m3-ln1) -- (m3-linear2) -- (m3-relu2) -- (m3-ln2) -- (m3-linear3) -- (m3-resplus);

        \node (m3-res1) [inner sep=0] at ($ (m3-dummy1 |- m3-ln1.north west) + (0, 0.25*\unit) $) {};
        \node (m3-res2) [inner sep=0] at ($ (m3-resplus |- m3-linear3.north east) + (0, 0.25*\unit) $) {};
        \draw [curved-line] (m3-dummy1.center) -- (m3-res1.center) -- (m3-res2.center) -- (m3-resplus) -- (m3-position);

        \draw [arrow] (m3-position) -- (m3-h);

        % Reasoner
        \begin{scope}[node distance = 0.1*\unit]
            \node (r-hi) [empty-rotated-embedding, minimum height=0.2*\unit, above=of m3-h] {\rvdots};
            \node (r-h1) [rotated-embedding, above=of r-hi] {};
            \node (r-hj) [empty-rotated-embedding, minimum height=0.2*\unit, below=of m3-h] {\rvdots};
            \node (r-h9) [rotated-embedding, below=of r-hj] {};
        \end{scope}
        \node [rotated-label] at (r-h1) {$h_1$};
        \node [rotated-label] at (r-h9) {$h_k$};

        \draw [arrow] ($ (m3-position.east |- r-h1) $) -- (r-h1);
        \draw [arrow] ($ (m3-position.east |- r-h9) $) -- (r-h9);

        \begin{scope}[node distance = 0.05*\unit]
            \node (r-stack) [rotated-layer, right=of m3-h, xshift=0.2*\unit] {};
            \node (r-bn1) [rotated-layer, norm, right=of r-stack] {};
            \node (r-pathways1) [rotated-layer, pathways, right=of r-bn1] {};
            \node (r-bottleneck1) [rotated-layer, right=of r-pathways1, xshift=0.1*\unit] {};
            \node (r-bn2) [rotated-layer, norm, right=of r-bottleneck1] {};
            \node (r-pathways2) [rotated-layer, pathways, right=of r-bn2] {};
            \node (r-adaptive-avgpool) [rotated-layer, right=of r-pathways2, xshift=0.1*\unit] {};
            \node (r-flatten) [rotated-layer, right=of r-adaptive-avgpool] {};
            \node (r-linear1) [rotated-layer, linear, right=of r-flatten] {};
            \node (r-relu1) [rotated-layer, activation, right=of r-linear1] {};
            \node (r-bn3) [rotated-layer, norm, right=of r-relu1] {};
            \node (r-linear2) [rotated-layer, linear, right=of r-bn3] {};
            \node (r-z) [rotated-embedding, right=of r-linear2, xshift=0.2*\unit] {};
        \end{scope}
        \node [rotated-label] at (r-stack) {Stack};
        \node [rotated-label] at (r-bn1) {BN1D};
        \node [rotated-label] at (r-pathways1) {Pathways};
        \node [rotated-label] at (r-bottleneck1) {B-Neck};
        \node [rotated-label] at (r-bn2) {BN1D};
        \node [rotated-label] at (r-pathways2) {Pathways};
        \node [rotated-label] at (r-adaptive-avgpool) {A-AvgPool};
        \node [rotated-label] at (r-flatten) {Flatten};
        \node [rotated-label] at (r-linear1) {Linear};
        \node [rotated-label] at (r-relu1) {ReLU};
        \node [rotated-label] at (r-bn3) {BN1D};
        \node [rotated-label] at (r-linear2) {Linear};
        \node [rotated-label] at (r-z) {$z_k$};

        \begin{scope}[on background layer]
            \node (r-bg) [fit={(r-bottleneck1) (r-pathways2)}, background] {};
        \end{scope}
        \node (r-2x) [text-label, anchor=north west] at (r-bg.south west) {$2 \times$};

        \draw (m3-h) -- (r-stack) -- (r-bn1) -- (r-pathways1) -- (r-bottleneck1) -- (r-bn2) -- (r-pathways2) -- (r-adaptive-avgpool) -- (r-flatten) -- (r-linear1) -- (r-relu1) -- (r-bn3) -- (r-linear2) [arrow] -- (r-z);

        \draw [curved-line] (r-h1.east) to[out=330, in=150] ($ (r-stack.north west)!1/4!(r-stack.south west) $);
        \draw [curved-line] (r-h9.east) to[out=30, in=210] ($ (r-stack.north west)!3/4!(r-stack.south west) $);

        \node (caption-a-x) [inner sep=0] at ($ (panel)!0.5!(r-z) $) {};
        \node (caption-a) [caption] at ($ (caption-a-x |- m12-bg.south) $) {(a) End-to-end PoNG architecture};

        % === Pathways ===

        % Pathways input
        \node (pathways-input) [inner sep=0, below=of panel, xshift=-0.375*\unit, yshift=-0.9*\unit] {
            \begin{tikzpicture}
                \foreach \i in {1,...,9} {
                    \foreach \j in {1,...,3} {
                        \pgfmathtruncatemacro{\colorindex}{int((\i - 1) * 3 + \j)}
                        \rec[]{\i}{\j}{spectral-\colorindex}
                    }
                }
            \end{tikzpicture}
        };

        \node (pathways-12-start) [inner sep=0, xshift=0.35*\unit] at (pathways-input.east) {};

        % 1st pathway
        \begin{scope}[node distance = 0.1*\unit]
            \node (p1-conv) [rotated-layer, conv, minimum height=0.8*\unit, above=of pathways-12-start] {};
        \end{scope}
        \node [rotated-label] at (p1-conv) {Conv1D};

        % 2nd pathway
        \begin{scope}[node distance = 0.1*\unit]
            \node (p2-conv) [rotated-layer, conv, minimum height=0.8*\unit, below=of pathways-12-start] {};
            \begin{scope}[node distance = 0.05*\unit]
                \node (p2-relu) [rotated-layer, activation, minimum height=0.8*\unit, right=of p2-conv] {};
                \node (p2-bn) [rotated-layer, norm, minimum height=0.8*\unit, right=of p2-relu] {};
            \end{scope}
        \end{scope}
        \node [rotated-label] at (p2-conv) {Conv1D};
        \node [rotated-label] at (p2-relu) {ReLU};
        \node [rotated-label] at (p2-bn) {BN1D};
        \begin{scope}[on background layer]
            \node (p2-bg) [fit={(p2-conv) (p2-bn)}, background] {};
        \end{scope}
        \node (p2-2x) [rotated-text-label, anchor=north east] at (p2-bg.south west) {$2 \times$};

        \node (pathways-34-start) [inner sep=0, xshift=0.6*\unit] at ($ (p2-bn.east |- pathways-input) $) {};

        % 3rd pathway
        \begin{scope}[node distance = 0.1*\unit]
            \node (p3-conv) [rotated-layer, conv, minimum height=0.8*\unit, above=of pathways-34-start] {};
            \begin{scope}[node distance = 0.05*\unit]
                \node (p3-relu) [rotated-layer, activation, minimum height=0.8*\unit, right=of p3-conv] {};
                \node (p3-bn) [rotated-layer, norm, minimum height=0.8*\unit, right=of p3-relu] {};
            \end{scope}
        \end{scope}
        \node [rotated-label] at (p3-conv) {G-C};
        \node [rotated-label] at (p3-relu) {ReLU};
        \node [rotated-label] at (p3-bn) {BN1D};
        \begin{scope}[on background layer]
            \node (p3-bg) [fit={(p3-conv) (p3-bn)}, background] {};
        \end{scope}
        \node (p3-2x) [rotated-text-label, anchor=north west] at (p3-bg.north west) {$2 \times$};

        % 4th pathway
        \begin{scope}[node distance = 0.1*\unit]
            \node (p4-conv) [rotated-layer, conv, minimum height=0.8*\unit, below=of pathways-34-start] {};
            \begin{scope}[node distance = 0.05*\unit]
                \node (p4-relu) [rotated-layer, activation, minimum height=0.8*\unit, right=of p4-conv] {};
                \node (p4-bn) [rotated-layer, norm, minimum height=0.8*\unit, right=of p4-relu] {};
            \end{scope}
        \end{scope}
        \node [rotated-label] at (p4-conv) {GP-C};
        \node [rotated-label] at (p4-relu) {ReLU};
        \node [rotated-label] at (p4-bn) {BN1D};
        \begin{scope}[on background layer]
            \node (p4-bg) [fit={(p4-conv) (p4-bn)}, background] {};
        \end{scope}
        \node (p4-2x) [rotated-text-label, anchor=north east] at (p4-bg.south west) {$2 \times$};

        % Pathways output

        \node (pathways-resplus) [resplus, xshift=0.35*\unit] at ($ (p4-bn.east |- pathways-input) $) {$+$};
        \begin{scope}[node distance = 0.25*\unit]
            \node (pathways-output) [inner sep=0, right=of pathways-resplus] {
                \begin{tikzpicture}
                    \foreach \i in {1,...,9} {
                        \foreach \j in {1,...,3} {
                            \pgfmathtruncatemacro{\colorindex}{int((\i - 1) * 3 + \j)}
                            \rec[]{\i}{\j}{spectral-\colorindex}
                        }
                    }
                \end{tikzpicture}
            };
        \end{scope}

        % Pathways lines
        \node (p1-l) [inner sep=0, yshift=4pt] at (p3-2x.west) {};
        \draw [curved-line] (pathways-input.east) to[out=0, in=180] (p1-conv) to[out=0, in=180] (p1-l.center) -- ($ (pathways-resplus |- p1-l) $) -- (pathways-resplus);

        \node (p2-l) [inner sep=0, yshift=-4pt] at (p4-2x.east) {};
        \draw [curved-line] (pathways-input.east) to[out=0, in=180] (p2-bg);
        \draw [curved-line] (p2-bg.east) to[out=330, in=180] (p2-l.center) -- ($ (pathways-resplus |- p2-l) $) -- (pathways-resplus);

        \node (p34-l) [inner sep=0] at ($ (p2-bg.east |- pathways-input) $) {};
        \draw (pathways-input.east) -- (p34-l.center) to[out=0, in=180] (p3-bg) to[out=0, in=135] (pathways-resplus.north west);
        \draw (pathways-input.east) -- (p34-l.center) to[out=0, in=180] (p4-bg) to[out=0, in=225] (pathways-resplus.south west);

        \draw [arrow] (pathways-resplus) -- (pathways-output);

        \draw (p2-conv) -- (p2-relu) -- (p2-bn);
        \draw (p3-conv) -- (p3-relu) -- (p3-bn);
        \draw (p4-conv) -- (p4-relu) -- (p4-bn);

        \node (caption-b-x) [inner sep=0] at ($ (pathways-input)!0.5!(pathways-output) $) {};
        \node (caption-b) [caption, yshift=-0.2*\unit] at ($ (caption-b-x |- p2-l.south) $) {(b) Pathways block};

        % === SharedGroupConv1d ===

        % Input matrix
        \node (input) [inner sep=0] at ($ (pathways-output) + (1.25*\unit, 0) $) {
            \begin{tikzpicture}
                \foreach \i in {1,...,9} {
                    \foreach \j in {1,...,3} {
                        \pgfmathtruncatemacro{\colorindex}{int((\i - 1) * 3 + \j)}
                        \rec[]{\i}{\j}{spectral-\colorindex}
                    }
                }
            \end{tikzpicture}
        };

        % Input groups
        \begin{scope}[node distance = 0.25*\unit]
            \node (input-group-2) [inner sep=0, right=of input, xshift=-0.1*\unit] {
                \begin{tikzpicture}
                    \foreach \i in {4,...,6} {
                        \foreach \j in {1,...,3} {
                            \pgfmathtruncatemacro{\colorindex}{int((\i - 1) * 3 + \j)}
                            \rec[]{\i}{\j}{spectral-\colorindex}
                        }
                    }
                \end{tikzpicture}
            };
            \node (input-group-1) [inner sep=0, above=of input-group-2] {
                \begin{tikzpicture}
                    \foreach \i in {1,...,3} {
                        \foreach \j in {1,...,3} {
                            \pgfmathtruncatemacro{\colorindex}{int((\i - 1) * 3 + \j)}
                            \rec[]{\i}{\j}{spectral-\colorindex}
                        }
                    }
                \end{tikzpicture}
            };
            \node (input-group-3) [inner sep=0, below=of input-group-2] {
                \begin{tikzpicture}
                    \foreach \i in {7,...,9} {
                        \foreach \j in {1,...,3} {
                            \pgfmathtruncatemacro{\colorindex}{int((\i - 1) * 3 + \j)}
                            \rec[]{\i}{\j}{spectral-\colorindex}
                        }
                    }
                \end{tikzpicture}
            };
        \end{scope}

        % Layers: Vertical variant
        \begin{scope}[node distance = 0.25*\unit]
            \node (group-conv-1) [rotated-layer, conv, minimum height=0.8*\unit, right=of input-group-1] {};
            \node (group-conv-2) [empty-rotated-layer, minimum height=0.8*\unit, right=of input-group-2] {};
            \node (group-conv-3) [rotated-layer, conv, minimum height=0.8*\unit, right=of input-group-3] {};
        \end{scope}
        \begin{scope}[node distance = 0.1*\unit]
            \node (norm-1) [rotated-layer, norm, minimum height=0.8*\unit, right=of group-conv-1] {};
            \node (norm-2) [empty-rotated-layer, minimum height=0.8*\unit, right=of group-conv-2] {};
            \node (norm-3) [rotated-layer, norm, minimum height=0.8*\unit, right=of group-conv-3] {};
        \end{scope}
        \node [rotated-label] at (group-conv-1) {Conv1D};
        \node [label] at (group-conv-2) {\rvdots};
        \node [rotated-label] at (group-conv-3) {Conv1D};
        \node [rotated-label] at (norm-1) {TCN};
        \node [label] at (norm-2) {\rvdots};
        \node [rotated-label] at (norm-3) {TCN};
        
        % Layers: Horizontal variant
        % \begin{scope}[node distance = 0.25*\unit]
        %     \node (group-conv-1) [layer, right=of input-group-1] {Conv1D};
        %     \node (group-conv-2) [layer, right=of input-group-2] {Conv1D};
        %     \node (group-conv-3) [layer, right=of input-group-3] {Conv1D};

        %     \node (norm-1) [layer, minimum width=0.6*\unit, right=of group-conv-1] {TCN};
        %     \node (norm-2) [layer, minimum width=0.6*\unit, right=of group-conv-2] {TCN};
        %     \node (norm-3) [layer, minimum width=0.6*\unit, right=of group-conv-3] {TCN};
        % \end{scope}

        % Barrier
        \node (barrier-top-x) [inner sep=0] at ($ (input-group-1.east)!0.5!(group-conv-1.west) $) {};
        \node (barrier-bot-x) [inner sep=0] at ($ (input-group-3.east)!0.5!(group-conv-3.west) $) {};
        \node (barrier-top) [inner sep=0] at ($ (barrier-top-x |- group-conv-1.north) $) {};
        \node (barrier-bot) [inner sep=0] at ($ (barrier-bot-x |- group-conv-3.south) $) {};
        \draw [red, dashed, very thick] (barrier-top.center) -- (barrier-bot.center);

        % Output groups
        \begin{scope}[node distance = 0.25*\unit]
            \node (output-group-1) [inner sep=0, right=of norm-1] {
                \begin{tikzpicture}
                    \foreach \i in {1,...,9} {
                        \foreach \j in {1,...,3} {
                            \pgfmathtruncatemacro{\colorindex}{int((\i - 1) * 3 + \j)}
                            \rec[]{\i}{\j}{spectral-1-\colorindex}
                        }
                    }
                \end{tikzpicture}
            };
            \node (output-group-3) [inner sep=0, right=of norm-3] {
                \begin{tikzpicture}
                    \foreach \i in {1,...,9} {
                        \foreach \j in {1,...,3} {
                            \pgfmathtruncatemacro{\colorindex}{int((\i - 1) * 3 + \j)}
                            \rec[]{\i}{\j}{spectral-3-\colorindex}
                        }
                    }
                \end{tikzpicture}
            };
            \node (output-group-2-dummy) [inner sep=0, xshift=0.1*\unit] at ($ (output-group-1.south)!0.5!(output-group-3.north) $) {};
            \node (output-group-2) [inner sep=0, right=of output-group-2-dummy] {
                \begin{tikzpicture}
                    \foreach \i in {1,...,9} {
                        \foreach \j in {1,...,3} {
                            \pgfmathtruncatemacro{\colorindex}{int((\i - 1) * 3 + \j)}
                            \rec[]{\i}{\j}{spectral-2-\colorindex}
                        }
                    }
                \end{tikzpicture}
            };
        \end{scope}

        % Output matrix
        \begin{scope}[node distance = 0.1*\unit]
            \node (output-resplus) [resplus, right=of output-group-2] {$+$};
        \end{scope}
        \begin{scope}[node distance = 0.25*\unit]
            \node (output) [inner sep=0, right=of output-resplus] {
                \begin{tikzpicture}
                    \foreach \i in {1,...,9} {
                        \foreach \j in {1,...,3} {
                            \pgfmathtruncatemacro{\colorindex}{int((\i - 1) * 3 + \j)}
                            \rec[]{\i}{\j}{spectral-\colorindex}
                        }
                    }
                \end{tikzpicture}
            };
        \end{scope}

        % Lines
        \draw ($ (input.north east)!3/18!(input.south east) $) to[out=0, in=180] (input-group-1.west);
        \draw (input) to[out=0, in=180] (input-group-2);
        \draw ($ (input.north east)!15/18!(input.south east) $) to[out=0, in=180] (input-group-3.west);

        \draw [arrow] (input-group-1) -- (group-conv-1) -- (norm-1) -- (output-group-1);
        \draw [arrow] (input-group-2) -- (group-conv-2) -- (norm-2) -- (output-group-2);
        \draw [arrow] (input-group-3) -- (group-conv-3) -- (norm-3) -- (output-group-3);

        \draw [curved-line] (output-group-1) -- ($ (output-resplus |- output-group-1) $) -- (output-resplus);
        \draw [arrow] (output-group-2) -- (output-resplus) -- (output);
        \draw [curved-line] (output-group-3) -- ($ (output-resplus |- output-group-3) $) -- (output-resplus);

        \node (caption-c-x) [inner sep=0] at ($ (input)!0.5!(output) $) {};
        \node (caption-c) [caption] at ($ (caption-c-x |- caption-b) $) {(c) Group convolution (P3)};

        % === GenericRowPairSharedGroupConv1d ===

        % Row-pair Input matrix
        \node (rowpair-input) [inner sep=0] at ($ (output) + (1.25*\unit, 0) $) {
            \begin{tikzpicture}
                \foreach \i in {1,...,9} {
                    \foreach \j in {1,...,3} {
                        \pgfmathtruncatemacro{\colorindex}{int((\i - 1) * 3 + \j)}
                        \rec[]{\i}{\j}{spectral-\colorindex}
                    }
                }
            \end{tikzpicture}
        };

        % Row-pair Input groups
        \begin{scope}[node distance = 0.25*\unit]
            \node (rowpair-input-group-2) [inner sep=0, right=of rowpair-input, xshift=-0.1*\unit] {
                \begin{tikzpicture}
                    \foreach \i in {4,...,6} {
                        \foreach \j in {1,...,3} {
                            \pgfmathtruncatemacro{\colorindex}{int((\i - 1) * 3 + \j)}
                            \rec[]{\i}{\j}{spectral-\colorindex}
                        }
                    }
                \end{tikzpicture}
            };
            \node (rowpair-input-group-1) [inner sep=0, above=of rowpair-input-group-2] {
                \begin{tikzpicture}
                    \foreach \i in {1,...,3} {
                        \foreach \j in {1,...,3} {
                            \pgfmathtruncatemacro{\colorindex}{int((\i - 1) * 3 + \j)}
                            \rec[]{\i}{\j}{spectral-\colorindex}
                        }
                    }
                \end{tikzpicture}
            };
            \node (rowpair-input-group-3) [inner sep=0, below=of rowpair-input-group-2] {
                \begin{tikzpicture}
                    \foreach \i in {7,...,9} {
                        \foreach \j in {1,...,3} {
                            \pgfmathtruncatemacro{\colorindex}{int((\i - 1) * 3 + \j)}
                            \rec[]{\i}{\j}{spectral-\colorindex}
                        }
                    }
                \end{tikzpicture}
            };
        \end{scope}

        % Row-pair Input group pairs
        \begin{scope}[node distance = 0.25*\unit]
            \node (rowpair-input-group-pair-1) [inner sep=0, right=of rowpair-input-group-1] {
                \begin{tikzpicture}
                    \foreach \i in {1,...,6} {
                        \foreach \j in {1,...,3} {
                            \pgfmathtruncatemacro{\colorindex}{int((\i - 1) * 3 + \j)}
                            \rec[]{\i}{\j}{spectral-\colorindex}
                        }
                    }
                \end{tikzpicture}
            };
            \node (rowpair-input-group-pair-3) [inner sep=0, right=of rowpair-input-group-3] {
                \begin{tikzpicture}
                    \foreach \i in {4,...,9} {
                        \foreach \j in {1,...,3} {
                            \pgfmathtruncatemacro{\colorindex}{int((\i - 1) * 3 + \j)}
                            \rec[]{\i}{\j}{spectral-\colorindex}
                        }
                    }
                \end{tikzpicture}
            };
            \node (rowpair-input-group-pair-2-dummy) [inner sep=0] at ($ (rowpair-input-group-pair-1.south)!0.5!(rowpair-input-group-pair-3.north) $) {};
            \node (rowpair-input-group-pair-2) [inner sep=0, right=of rowpair-input-group-pair-2-dummy, xshift=-0.1*\unit] {
                \begin{tikzpicture}
                    \foreach \i in {1,...,3} {
                        \foreach \j in {1,...,3} {
                            \pgfmathtruncatemacro{\colorindex}{int((\i - 1) * 3 + \j)}
                            \rec[]{\i}{\j}{spectral-\colorindex}
                        }
                    }
                    \foreach \i in {4,...,6} {
                        \foreach \j in {1,...,3} {
                            \pgfmathtruncatemacro{\colorindex}{int((\i - 1 + 3) * 3 + \j)}
                            \rec[]{\i}{\j}{spectral-\colorindex}
                        }
                    }
                \end{tikzpicture}
            };
            \node (rowpair-barrier-empty-layer) [empty-rotated-layer, right=of rowpair-input-group-pair-2] {};
        \end{scope}

        % Barrier
        \node (rowpair-barrier-x) [inner sep=0] at (rowpair-barrier-empty-layer.west) {};
        \node (rowpair-barrier-top) [inner sep=0] at ($ (rowpair-barrier-x |- rowpair-input-group-pair-1.north) $) {};
        \node (rowpair-barrier-bot) [inner sep=0] at ($ (rowpair-barrier-x |- rowpair-input-group-pair-3.south) $) {};
        \draw [red, dashed, very thick] (rowpair-barrier-top.center) -- (rowpair-barrier-bot.center);

        % Lines
        \draw ($ (rowpair-input.north east)!3/18!(rowpair-input.south east) $) to[out=0, in=180] (rowpair-input-group-1.west);
        \draw (rowpair-input) -- (rowpair-input-group-2);
        \draw ($ (rowpair-input.north east)!15/18!(rowpair-input.south east) $) to[out=0, in=180] (rowpair-input-group-3.west);

        \draw (rowpair-input-group-1.east) to[out=0, in=180] ($ (rowpair-input-group-pair-1.north west)!3/12!(rowpair-input-group-pair-1.south west) $);
        \draw (rowpair-input-group-2.east) to[out=0, in=180] ($ (rowpair-input-group-pair-1.north west)!9/12!(rowpair-input-group-pair-1.south west) $);

        \node (rpigp21) [inner sep=0] at ($ (rowpair-input-group-pair-2.north west)!3/12!(rowpair-input-group-pair-2.south west) $) {};
        \node (rpigp23) [inner sep=0] at ($ (rowpair-input-group-pair-2.north west)!9/12!(rowpair-input-group-pair-2.south west) $) {};
        \draw (rowpair-input-group-1.east) to[out=0, in=180] ($ (rowpair-input-group-pair-1.south west |- rpigp21) $) -- (rpigp21.center);
        \draw (rowpair-input-group-3.east) to[out=0, in=180] ($ (rowpair-input-group-pair-3.north west |- rpigp23) $) -- (rpigp23.center);

        \draw (rowpair-input-group-2.east) to[out=0, in=180] ($ (rowpair-input-group-pair-3.north west)!3/12!(rowpair-input-group-pair-3.south west) $);
        \draw (rowpair-input-group-3.east) to[out=0, in=180] ($ (rowpair-input-group-pair-3.north west)!9/12!(rowpair-input-group-pair-3.south west) $);

        \draw [arrow] (rowpair-input-group-pair-1) -- ($ (rowpair-barrier-x |- rowpair-input-group-pair-1) $);
        \draw [arrow] (rowpair-input-group-pair-2) -- ($ (rowpair-barrier-x |- rowpair-input-group-pair-2) $);
        \draw [arrow] (rowpair-input-group-pair-3) -- ($ (rowpair-barrier-x |- rowpair-input-group-pair-3) $);

        \node (caption-d-x) [inner sep=0] at ($ (rowpair-input)!0.5!(rowpair-barrier-x) $) {};
        \node (caption-d) [caption] at ($ (caption-d-x |- caption-b) $) {(d) Group-pair convolution (P4)};

    \end{tikzpicture}
    \caption{\textbf{PoNG.}
    (a) The panel encoder embeds each input image $x_i$ independently, producing $h_i$.
    Context panel embeddings $\{h_i\}_{i=1}^8$ together with the embedding of $k$'th answer $h_k$ are stacked and processed with the reasoner, leading to $z_k$.
    (b) The pathways block, a key component of PoNG, comprises four parallel pathways P1 -- P4.
    (c) P3 and (d) P4 employ novel normalized group convolution operators.
    PosEmb denotes position embedding, G-C the group convolution module used in P3, and GP-C the group-pair convolution module used in P4.
    The red dashed line marks the point after which G-C and GP-C perform analogous computation.}
    \label{fig:model}
\end{figure*}

%% file: sections/method.tex
\section{Pathways of Normalized Group Convolution (PoNG) Model}
\label{sec:method}

We introduce PoNG (Fig.~\ref{fig:model}), a novel model that outcompetes baselines across a number of problem settings.
The model follows a typical two-stage design.
Firstly, it generates an embedding of each image panel.
Then, it aggregates representations of matrix panels to predict the index of the correct answer.
The details are described in
% CAMERA-READY VERSION
% \cite[Appendix~A]{malkinski2025advancing}.
% ARXIV VERSION
Appendix~\ref{sec:model-details}.

Let $(X, y, r)$ denote an AVR matrix, where $X = \{x_i\}_{i=1}^{n}$ is the set of $n$ image panels comprising $n_c$ context panels $\{x_i\}_{i=1}^{n_c}$ and $n_a$ answer panels $\{x_i\}_{i=n_c+1}^{n}$,
$x_i \in [0, 1]^{h \times w}, i=1,\ldots,n$ is a grayscale image of height $h$ and width $w$,
$y \in \{0, 1\}^{n_a}$ is the one-hot encoded index of the correct answer,
$r \in \{0, 1\}^{d_r}$ is the multi-hot encoded representation of matrix rules of dimensionality $d_r$ using sparse encoding~\cite{malkinski2020multi}.
%For RPMs $n_c=8$ and $n_a=8$, for the VAP dataset $n_c=5$ and $n_a=4$, and for VASR $n_c=3$ and $n_a=4$.
For RPMs $n_c=n_a=8$, for VAP $n_c=5, n_a=4$, and for VASR $n_c=3, n_a=4$.
In each experiment $h = w = 80$, while $d_r$ is determined by the number of different abstract structures in the corresponding dataset ($d_r=40$ for I-RAVEN and A-I-RAVEN, $d_r=48$ for I-RAVEN-Mesh, $d_r=50$ for PGM, and $d_r=28$ for VAP).
% RAVEN, A-I-RAVEN: 2 components, 4 rules, 5 attributes = 2 * 4 * 5 = 40
% I-RAVEN-Mesh: RAVEN; 4 rules, 2 attributes = 40 + 8 = 48
% PGM: 2 objects, 5 rules, 5 attributes = 2 * 5 * 5 = 50
% VAP: 7 domains, 4 rules = 7 * 4 = 28

\paragraph{Panel encoder.}
The first component of the model has the form $\mathcal{E} : x \to h$, where $h \in \mathbb{R}^{d_h}$ is the input panel embedding of dimensionality $d_h$.
Following RelBase~\cite{spratley2020closer}, the module comprises $2$ blocks of the same architecture.
Each block includes $2$ parallel pathways that build high-level and low-level features, resp.
The first one contains $2$ convolutional blocks, each with 2D convolution, ReLU, and Batch Normalization (BN)~\cite{ioffe2015batch}.
The second one contains 2D max pooling followed by 2D convolution.
The sum of both pathway results forms the block output.
Differently from RelBase, we flatten the height and width dimensions of the resultant embedding, pass it through a linear layer with ReLU, flatten the channel and spatial dimensions, and pass the tensor through a feed-forward residual block with Layer Normalization (LN)~\cite{ba2016layer}.
Finally, we concatenate the tensor with a position embedding (a learned $25$-dimensional vector for each panel in the context grid), leading to $h$.

\paragraph{Reasoner.}
The second component of the model has the form $\mathcal{R} : \{h_i\}_{i=1}^8 \cup h_k \to z_k$, where $h_k$ is the panel embedding of  $k$'th answer.
For each answer panel, the reasoner produces embedding $z_k$ that describes how well the considered answer fits into the matrix context.
% Firstly, the component concatenates a position embedding (a learned $25$ dimensional vector for each cell in the $3 \times 3$ contexxt grid) to the corresponding panel embedding.
Panel embeddings $\{h_i\}_{i=1}^8 \cup h_k$ are stacked and processed by a sequence of $3$ reasoning blocks interleaved with $2$ bottleneck layers for dimensionality reduction.
Each reasoning block comprises BN and $4$ parallel pathways, outputs of which are added together to form the output of the block.
Next, the latent representation is passed through adaptive average pooling, flattened, processed with a linear layer with ReLU, passed through BN and projected with a linear layer to $z_k \in \mathbb{R}^{128}$.

\paragraph{Pathways.}
The key aspect of the reasoner module are its pathways.
Each takes an input tensor of shape $(B, C, D)$, where $B$ is the batch size, $C$ is the number of channels, and $D$ is the feature dimension.
In the first reasoning block $D=d_h$ and $C$ corresponds to the number of panel embeddings in the considered group ($C=9$ for RPMs, $C=6$ for VAP, and $C=4$ for VASR).
Pathways are described as follows:
P1 -- a pointwise 1D convolution layer that mixes panel features at each spatial location;
P2 -- a sequence of 2 blocks, each comprising 1D convolution, ReLU, and BN, that builds higher level features spanning neighbouring spatial locations;
P3 -- analogous to P2, but 1D convolution is replaced with a group 1D convolution that splits the tensor into several groups along the channel dimension, applies a 1D convolution with shared weights to each group, and adds together the representations of each group;
P4 -- analogous to P3, but groups are arranged into pairs concatenated along the channel dimension and processed with a 1D convolution with shared weights.
In contrast to~\cite{krizhevsky2012imagenet}, the proposed group convolution layers in both P3 and P4 apply TCN~\cite{webb2020learning} to the outputs in each group.
In the first layer P3 and P4 split the input tensor into $3$ groups for RPMs and visual analogies, and into $2$ groups for VASR, which allows for producing embeddings of each matrix row and each pair of rows, resp.
Though we apply the pathways block in the AVR context, we envisage it as a generic module, also applicable to other settings involving a set of vector representations of shape $(B, C, D)$.

\paragraph{Answer prediction.}
Representations of the context matrix filled-in with the respective answer, $\{z_k\}_{k=1}^{n_a}$, are processed with three prediction heads.
The target head $\mathcal{P}^y : z_k \to \widehat{y_k}$ employs two linear layers interleaved with ReLU to produce score $\widehat{y_k} \in \mathbb{R}$ describing how well the answer $k$ aligns with the matrix context.
The aggregate rule head $\mathcal{P}^r_1 : \{z_k\}_{k=1}^{n_a} \to \widehat{r_1}$ computes the sum of inputs and processes it with two linear layers interleaved with ReLU, producing a latent prediction of matrix rules $\widehat{r_1} \in \mathbb{R}^{d_r}$.
We also introduce a novel target-conditioned rule head $\mathcal{P}^r_2 : \{z_k\}_{k=1}^{n_a} \to \widehat{r_2}$, which processes its input through a linear layer and computes a weighted sum of the resultant embeddings with weights given by the predicted probability distribution over the set of possible answers $\sigma(\{\widehat{y_k}\}_{k=1}^{n_a})$, where $\sigma$ denotes softmax.
The model is trained with a joint loss function $\mathcal{L} = \text{CE}(\sigma(\{\widehat{y_k}\}_{k=1}^{n_a}), y) + \beta \text{BCE}(\zeta(\widehat{r_1}, r)) + \gamma \text{BCE}(\zeta(\widehat{r_2}, r))$, where $\zeta$ denotes sigmoid, CE cross-entropy, BCE binary cross-entropy, $\beta=25$ and $\gamma=5$ are balancing coefficients.

%% file: sections/experiments.tex
\input{tables/single-task-learning}
\input{tables/stl-new-regimes}

\section{Experiments}
\label{sec:experiments}

We employ a set of diverse AVR tasks to evaluate PoNG's generalization capabilities.
Section~\ref{sec:experiments-tasks} introduces the selected datasets, Section~\ref{sec:experiments-setup} details the experimental setup, and Section~\ref{sec:experiments-results} presents the results.

\subsection{AVR datasets}
\label{sec:experiments-tasks}

AVR models are typically evaluated on RPM benchmarks, a problem set well-established in the literature. 
We utilize four RPM datasets: PGM~\cite{santoro2018measuring}, I-RAVEN~\cite{hu2021stratified}, I-RAVEN-Mesh
%\cite{malkinski2025airaven},
and A-I-RAVEN~\cite{malkinski2025airaven}.
We extend the evaluation of PoNG beyond RPMs, to two benchmarks comprising visual analogies with both synthetic~\cite{hill2018learning} and real-world~\cite{bitton2023vasr} images.

\paragraph{PGM.}
The PGM dataset was the first large-scale RPM benchmark designed to evaluate the AVR capabilities of deep learning models.
In PGM each matrix is defined by an abstract structure encompassing its rules, objects, and attributes.
To assess generalization, the dataset is divided into $8$ generalization regimes.
In the Neutral regime, the train, validation, and test splits share the same feature distribution, constituting an i.i.d. learning challenge.
In the remaining regimes, the train and validation splits share a common distribution, while the test split relies on a different distribution, enabling the evaluation of generalization to unseen feature combinations.
Each regime contains $1.42$M RPMs, where $1.2$M, $20$K, and $200$K belong to the train, validation, and test splits, resp.

\paragraph{I-RAVEN.}
The RAVEN dataset~\cite{zhang2019raven} was constructed to expand the range of visual configurations in RPMs.
It incorporates $7$ configurations that define object locations within the matrices.
For instance, in the \texttt{Left-Right} configuration, each panel is divided into left and right parts that can be governed by distinct rules.
A subsequent study identified a bias in the RAVEN's answer generation method, enabling models to learn shortcut solutions~\cite{hu2021stratified}.
To alleviate this issue, the I-RAVEN dataset was proposed, which employs an impartial answer generation method.
We utilize I-RAVEN in the experiments to avoid learning shortcut solutions.
The benchmark consists of $10$K matrices per configuration, totaling $70$K matrices, split into the train, validation, and test with a $60/20/20$ ratio.

\paragraph{I-RAVEN-Mesh.}
The I-RAVEN-Mesh dataset builds upon I-RAVEN by rendering a grid of $1$ to $12$ lines on the underlying matrices.
The grid is defined by two attributes, the number of lines and their position.
While the dataset was originally introduced to assess knowledge acquisition in transfer learning settings, we use it for standard supervised learning.
In this setup, the model is trained directly on the dataset, analogously to I-RAVEN, to expand the scope of the considered i.i.d. tasks.

\paragraph{A-I-RAVEN.}
The A-I-RAVEN dataset was introduced to combine the generalization assessment capabilities of PGM with the broad adoption of RAVEN-like benchmarks.
Drawing from PGM, A-I-RAVEN defines $10$ generalization regimes.
In each regime, a subset of attributes follows specific rules in the train and validation splits, while being governed by different rules in the test split.
For example, in the \texttt{A/ColorSize} regime, the Color and Size attributes adhere to the Constant rule in the train and validation splits and are governed by a rule other than Constant in the test split.
This approach enables the evaluation of models on RPMs with novel rule--attribute combinations that were not seen during training.
Each regime contains $70$K matrices, analogously to I-RAVEN.

\input{tables/pgm}
\input{tables/vap}

\paragraph{VAP.}
The VAP benchmark was introduced to assess the analogy-making capabilities of learning systems.
Each VAP matrix consists of a $2\times3$ grid of panels.
The task is to identify a concept in the source domain (top row) and instantiate it in the target domain (bottom row) by selecting the correct answer panel to complete the matrix.
The dataset defines five generalization regimes: Novel Domain Transfer, Novel Target Domain: Colour of Shapes, Novel Target Domain: Type of Lines, Novel Attribute Values: Interpolation, and Novel Attribute Values: Extrapolation, which test the model's generalization to novel domains or attribute values.
Each regime contains $710$K matrices, with $600$K, $10$K, and $100$K devoted to the train, validation, and test splits, resp.
In all experiments we use the learning analogies by contrasting (LABC) dataset variant, which constructs the answer set using semantically plausible images that consistently complete the target domain with some relation.

\paragraph{VASR.}
The VASR dataset features visual analogies involving real-world images, requiring the learner to understand complex real-world scenes before solving the analogy problem.
Each matrix consists of a $2\times2$ panel grid, with the bottom-right image missing.
The task is to complete the matrix by selecting the correct image from the $4$ provided choices.
VASR follows the classical analogy problem formulation, which aims to complete the following relation: A is to B, as C is to D.
We use Silver data for training, which includes $150$K, $2.25$K, and $2.55$K matrices in the train, validation, and test splits, resp.
Experiments are conducted on both dataset variants, featuring random and difficult distractors, resp.

\subsection{Experimental setting}
\label{sec:experiments-setup}

We assess PoNG's generalization by comparing its performance to SOTA models on the respective datasets.
PoNG is trained using a standard training strategy involving the Adam optimizer~\cite{kingma2014adam} with default hyperparameters ($\lambda = 0.001$, $\beta_1=0.9$, $\beta_2=0.999$, $\epsilon=10^{-8}$).
Learning rate $\lambda$ is reduced by a factor of $10$ after $5$ epochs without improvement in validation loss.
Early stopping is applied after $10$ epochs without validation loss reduction.
We use batch size $B = 128$ for experiments on RAVEN-like datasets and $B = 256$ in the remaining cases to reduce training time on large datasets.
All experiments are performed on a single GPU (NVIDIA DGX A100).

To ensure reproducibility, we use a set of fixed random seeds, provide a list of commands for running training jobs, and explicitly list static dependencies in configuration files.
The code for reproducing all experiments is publicly accessible at: \url{https://github.com/mikomel/raven}

\subsection{Results}
\label{sec:experiments-results}

\paragraph{Results on I-RAVEN and I-RAVEN-Mesh.}
We begin with evaluating PoNG in the i.i.d. setting on the I-RAVEN and I-RAVEN-Mesh datasets, comparing it to $13$ SOTA baselines.
As presented in Table~\ref{tab:attributeless}, on I-RAVEN, using our experimental setup, PoNG achieves a test accuracy of $95.9\%$, outperforming all other models.
When compared to results obtained with model-specific experimental setups (I-RAVEN$^\dagger$), PoNG is placed just behind CPCNet, DRNet, and PredRNet, which achieve slightly better scores.
PoNG also secures the 1st place on I-RAVEN-Mesh, demonstrating high capacity to handle matrices with rules that span a large number of objects.
Unlike many baseline models that rely on deeper architectures such as DRNet, SRAN or STSN, PoNG presents competitive performance despite its parameter-efficient design.
% The negligible performance difference between the validation and test splits confirms that I-RAVEN's training and testing distributions are consistent.
These results demonstrate PoNG's strong ability to solve i.i.d. RPM tasks.

\paragraph{Results on A-I-RAVEN.}
To assess generalization, we evaluate PoNG on the $4$ primary regimes of the A-I-RAVEN dataset, where the training and test distributions differ significantly.
As shown in Table~\ref{tab:attributeless}, PoNG outperforms all baselines across all settings, achieving test accuracies ranging from $59.4\%$ on \texttt{A/Type} to $80.3\%$ on \texttt{A/Color}, surpassing the best reference models by $10.3$ and $9.9$ p.p., resp.
Additionally, Table~\ref{tab:attributeless-extended} shows PoNG's performance across $6$ extended regimes, which cover more challenging generalization tasks.
Similarly, PoNG achieves superior performance in all but one regimes.
Notably, PoNG outperforms the 2nd best model in the \texttt{A/Color-D3} regime by $15.6$ p.p.
Overall, the results on A-I-RAVEN highlight PoNG's ability to perform well across a wide range of generalization tasks with varying levels of complexity.
However, certain regimes such as the $3$ extended regimes with held-out attribute pairs (\texttt{A/ColorSize}, \texttt{A/ColorType}, \texttt{A/SizeType}) continue to pose a significant challenge for all models (including PoNG), raising the need for further advances in generalization.

\input{tables/vasr}

\input{tables/ablations}

\paragraph{Results on PGM.}
Table~\ref{tab:pgm} presents PoNG's results across PGM regimes.
The model achieves strong results in several settings, particularly excelling in the Held-out Triple Pairs regime, where it surpasses the best reference model by $19.4$ p.p.
On average, PoNG scored $57.3\%$ accuracy securing the 2nd place, just behind DRNet with $58.3\%$.
These results confirm PoNG's ability to perform well on RPM-based generalization challenges extending beyond the RAVEN dataset line.

\paragraph{Synthetic visual analogies.}
Table~\ref{tab:vap} presents PoNG's results across $5$ regimes from the VAP benchmark.
PoNG achieves SOTA results in $3$ out of $5$ settings when compared to PredRNet, the currently leading VAP model.
The Novel Attribute Values: Extrapolation regime poses the greatest challenge among VAP regimes, aligning with findings from PGM, where Extrapolation is also one of the most demanding regimes.
Overall, PoNG and PredRNet perform competitively, with PredRNet achieving a better average score by $1$ p.p.
PoNG’s strong results on VAP highlight its versatility in generalization tasks that extend beyond RPMs.

\paragraph{Real-World visual analogies.}
To evaluate PoNG on the VASR dataset we followed the approach proposed by the VASR authors and employed the Vision Transformer (ViT)~\cite{dosovitskiy2021an} as a perception backbone that produces image embeddings.
Specifically, we used the same model variant as~\cite{bitton2023vasr}, which is ViT-L/32 pre-trained on ImageNet-21k at resolution 224x224 and fine-tuned on ImageNet-1k at resolution 384x384.
We replaced the panel encoder of PoNG with this frozen pre-trained backbone and trained the rest of the model from scratch.
The results are presented in Table~\ref{tab:vasr}.
The three reference methods perform comparably to each other, with Supervised Concat being slightly inferior to Zero-Shot methods on the random distractor split and slightly superior on the difficult split.
However, in both dataset variants PoNG significantly outcompetes the strongest reference result with $92.0\%$ vs. $86.0\%$ and $70.5\%$ vs. $54.9\%$, resp.
This suggests that the proposed reasoner block is much more effective in reasoning over pre-trained embeddings than baseline methods.
The results support the claim that PoNG is a versatile model with strong analogical reasoning capabilities, applicable to both synthetic and real-world domains.

\paragraph{Ablation study.}
% 1) each pathways block was simplified by removing P3 and P4 (w/o P3 and P4);
% 2) without $\mathcal{P}_2^r$ ($\gamma=0$);
% 3) without $\mathcal{P}_1^r$ ($\beta=0$);
% 4) without $\mathcal{P}_2^r$ and $\mathcal{P}_1^r$ ($\gamma=0 \land \beta=0$);
% 5) without TCN (w/o TCN);
% 6) all of the above.
We performed an ablation study on the RAVEN dataset line to evaluate the contributions of different PoNG components.
Table~\ref{tab:ablations} summarizes the results.
The removal of P1 and P2 (cf. Fig.~\ref{fig:model}) leads to performance drop, in particular on I-RAVEN-Mesh ($-14.9$ p.p.) and \texttt{A/Size} ($-15.2$ p.p.).
Similarly, removing P3 and P4 reduces model performance, especially on \texttt{A/Type} ($-5.5$ p.p.).
Disabling TCN leads to generally worse results, primarily on \texttt{A/Color} ($-4.9$ p.p.) and \texttt{A/Size} ($-6.9$ p.p.).
As shown in
% CAMERA-READY VERSION
% \cite[Appendix~B]{malkinski2025advancing},
% ARXIV VERSION
Appendix~\ref{sec:qualitative-analysis},
PoNG w/o TCN may fail to generalize rules to held-out attributes.
Training without $\mathcal{P}_1^r$ ($\beta=0$) or $\mathcal{P}_2^r$ ($\gamma=0$) typically reduces model performance, but training with one of these rule-based prediction heads compensates to some degree the lack of the other.
However, the removal of both ($\gamma=0 \land \beta=0$) deteriorates results across all datasets, signifying high relevance of the auxiliary training signal in PoNG's training.
Overall, the ablation study demonstrates that all employed design choices contribute to the model performance.

%% file: tables/single-task-learning.tex
% \addtolength{\tabcolsep}{-3pt}
\begin{table*}[t]
\centering
\small
\begin{tabular}{lccccccc}
\toprule
 & \multicolumn{3}{c}{i.i.d. tasks: I-RAVEN and I-RAVEN-Mesh} & \multicolumn{4}{c}{o.o.d. tasks: A-I-RAVEN} \\
 \cmidrule(lr){2-4} \cmidrule(lr){5-8}
 & I-RAVEN$^\dagger$ & I-RAVEN & I-RAVEN-Mesh & \texttt{A/Color} & \texttt{A/Position} & \texttt{A/Size} & \texttt{A/Type} \\
\midrule
ALANS & $-$ & $27.0$ \scriptsize$(\pm\,8.4)$ & $15.9$ \scriptsize$(\pm\,2.6)$ & $15.2$ \scriptsize$(\pm\,1.4)$ & $16.0$ \scriptsize$(\pm\,1.0)$ & $23.3$ \scriptsize$(\pm\,6.5)$ & $19.0$ \scriptsize$(\pm\,3.4)$ \\
CPCNet & $\textbf{98.5}$ & $70.4$ \scriptsize$(\pm\,6.4)$ & $66.6$ \scriptsize$(\pm\,5.1)$ & $51.2$ \scriptsize$(\pm\,3.8)$ & $68.3$ \scriptsize$(\pm\,4.0)$ & $43.5$ \scriptsize$(\pm\,3.5)$ & $38.6$ \scriptsize$(\pm\,4.3)$ \\
CNN-LSTM & $18.9$ & $27.5$ \scriptsize$(\pm\,1.5)$ & $28.9$ \scriptsize$(\pm\,0.4)$ & $17.0$ \scriptsize$(\pm\,3.1)$ & $24.0$ \scriptsize$(\pm\,2.9)$ & $13.6$ \scriptsize$(\pm\,1.4)$ & $14.5$ \scriptsize$(\pm\,0.8)$ \\
CoPINet & $46.1$ & $43.2$ \scriptsize$(\pm\,0.1)$ & $41.1$ \scriptsize$(\pm\,0.3)$ & $32.5$ \scriptsize$(\pm\,0.2)$ & $41.3$ \scriptsize$(\pm\,1.6)$ & $21.8$ \scriptsize$(\pm\,0.2)$ & $19.8$ \scriptsize$(\pm\,0.9)$ \\
DRNet & $\underline{97.6}$ & $\underline{90.9}$ \scriptsize$(\pm\,1.1)$ & $83.9$ \scriptsize$(\pm\,2.7)$ & $\underline{70.0}$ \scriptsize$(\pm\,1.6)$ & $\underline{77.5}$ \scriptsize$(\pm\,0.9)$ & $54.3$ \scriptsize$(\pm\,3.0)$ & $44.3$ \scriptsize$(\pm\,0.8)$ \\
MRNet & $83.5$ & $86.7$ \scriptsize$(\pm\,2.3)$ & $79.5$ \scriptsize$(\pm\,2.0)$ & $33.6$ \scriptsize$(\pm\,8.2)$ & $62.6$ \scriptsize$(\pm\,2.6)$ & $20.6$ \scriptsize$(\pm\,5.0)$ & $19.4$ \scriptsize$(\pm\,0.3)$ \\
PrAE & $77.0$ & $19.5$ \scriptsize$(\pm\,0.4)$ & $33.2$ \scriptsize$(\pm\,0.4)$ & $47.9$ \scriptsize$(\pm\,0.9)$ & $68.2$ \scriptsize$(\pm\,3.3)$ & $41.3$ \scriptsize$(\pm\,1.8)$ & $37.0$ \scriptsize$(\pm\,1.7)$ \\
PredRNet & $96.5$ & $88.8$ \scriptsize$(\pm\,1.8)$ & $59.2$ \scriptsize$(\pm\,6.4)$ & $59.4$ \scriptsize$(\pm\,1.0)$ & $73.7$ \scriptsize$(\pm\,0.7)$ & $47.5$ \scriptsize$(\pm\,1.3)$ & $40.2$ \scriptsize$(\pm\,1.3)$ \\
RelBase & $91.1$ & $89.6$ \scriptsize$(\pm\,0.6)$ & $\underline{84.9}$ \scriptsize$(\pm\,4.4)$ & $67.4$ \scriptsize$(\pm\,2.7)$ & $76.6$ \scriptsize$(\pm\,0.3)$ & $51.1$ \scriptsize$(\pm\,2.4)$ & $44.1$ \scriptsize$(\pm\,1.0)$ \\
SCL & $95.0$ & $83.4$ \scriptsize$(\pm\,2.5)$ & $80.9$ \scriptsize$(\pm\,1.5)$ & $65.1$ \scriptsize$(\pm\,2.0)$ & $76.7$ \scriptsize$(\pm\,7.1)$ & $\underline{65.6}$ \scriptsize$(\pm\,2.4)$ & $\underline{49.5}$ \scriptsize$(\pm\,1.8)$ \\
SRAN & $60.8$ & $58.2$ \scriptsize$(\pm\,1.6)$ & $57.8$ \scriptsize$(\pm\,0.2)$ & $38.3$ \scriptsize$(\pm\,1.0)$ & $56.9$ \scriptsize$(\pm\,0.7)$ & $34.4$ \scriptsize$(\pm\,3.0)$ & $30.7$ \scriptsize$(\pm\,2.2)$ \\
STSN & $95.7$ & $51.0$ \scriptsize$(\pm\,24.8)$ & $48.7$ \scriptsize$(\pm\,11.5)$ & $39.3$ \scriptsize$(\pm\,6.9)$ & $36.1$ \scriptsize$(\pm\,19.9)$ & $38.4$ \scriptsize$(\pm\,16.6)$ & $39.1$ \scriptsize$(\pm\,5.0)$ \\
WReN & $23.8$ & $18.4$ \scriptsize$(\pm\,0.0)$ & $25.7$ \scriptsize$(\pm\,0.2)$ & $16.9$ \scriptsize$(\pm\,0.5)$ & $17.3$ \scriptsize$(\pm\,0.4)$ & $12.4$ \scriptsize$(\pm\,0.5)$ & $15.1$ \scriptsize$(\pm\,0.7)$ \\
\midrule
PoNG (ours) & $95.9$ & $\textbf{95.9}$ \scriptsize$(\pm\,0.7)$ & $\textbf{89.3}$ \scriptsize$(\pm\,2.4)$ & $\textbf{80.3}$ \scriptsize$(\pm\,4.3)$ & $\textbf{79.3}$ \scriptsize$(\pm\,0.7)$ & $\textbf{73.5}$ \scriptsize$(\pm\,3.1)$ & $\textbf{59.4}$ \scriptsize$(\pm\,6.9)$ \\
\bottomrule
\end{tabular}
\caption{\textbf{RAVEN-related datasets.} Mean and standard deviation of test accuracy for three random seeds. Best dataset results are marked in bold and the second best are underlined. I-RAVEN$^\dagger$ denotes results on I-RAVEN reported by model authors in the corresponding papers.}
\label{tab:attributeless}
\end{table*}
% \addtolength{\tabcolsep}{3pt}

%% file: tables/stl-new-regimes.tex
% \addtolength{\tabcolsep}{-1pt}
\begin{table*}[t]
\centering
\small
\begin{tabular}{lcccccc}
\toprule
 & \texttt{A/ColorSize} & \texttt{A/ColorType} & \texttt{A/SizeType} & \texttt{A/Color-P} & \texttt{A/Color-A} & \texttt{A/Color-D3} \\
\midrule
ALANS & $15.1$ \scriptsize$(\pm\,3.3)$ & $17.7$ \scriptsize$(\pm\,3.2)$ & $15.7$ \scriptsize$(\pm\,3.2)$ & $24.8$ \scriptsize$(\pm\,18.8)$ & $18.3$ \scriptsize$(\pm\,6.6)$ & $22.4$ \scriptsize$(\pm\,7.7)$ \\
CPCNet & $33.0$ \scriptsize$(\pm\,5.3)$ & $25.0$ \scriptsize$(\pm\,0.9)$ & $24.1$ \scriptsize$(\pm\,1.2)$ & $50.5$ \scriptsize$(\pm\,0.6)$ & $45.9$ \scriptsize$(\pm\,2.7)$ & $37.8$ \scriptsize$(\pm\,0.9)$ \\
CNN-LSTM & $13.4$ \scriptsize$(\pm\,0.9)$ & $14.7$ \scriptsize$(\pm\,1.7)$ & $13.0$ \scriptsize$(\pm\,0.1)$ & $17.2$ \scriptsize$(\pm\,1.5)$ & $17.1$ \scriptsize$(\pm\,3.7)$ & $20.6$ \scriptsize$(\pm\,6.7)$ \\
CoPINet & $18.3$ \scriptsize$(\pm\,0.3)$ & $17.2$ \scriptsize$(\pm\,0.1)$ & $19.7$ \scriptsize$(\pm\,0.7)$ & $35.8$ \scriptsize$(\pm\,0.6)$ & $35.2$ \scriptsize$(\pm\,0.5)$ & $26.9$ \scriptsize$(\pm\,0.5)$ \\
DRNet & $38.3$ \scriptsize$(\pm\,0.5)$ & $29.5$ \scriptsize$(\pm\,0.5)$ & $31.6$ \scriptsize$(\pm\,1.2)$ & $72.8$ \scriptsize$(\pm\,1.3)$ & $\underline{66.7}$ \scriptsize$(\pm\,1.2)$ & $63.2$ \scriptsize$(\pm\,0.3)$ \\
MRNet & $18.7$ \scriptsize$(\pm\,1.1)$ & $20.0$ \scriptsize$(\pm\,2.6)$ & $28.2$ \scriptsize$(\pm\,0.9)$ & $34.4$ \scriptsize$(\pm\,3.4)$ & $35.7$ \scriptsize$(\pm\,5.9)$ & $18.6$ \scriptsize$(\pm\,0.1)$ \\
PrAE & $30.0$ \scriptsize$(\pm\,1.1)$ & $26.7$ \scriptsize$(\pm\,0.7)$ & $25.6$ \scriptsize$(\pm\,0.8)$ & $62.3$ \scriptsize$(\pm\,0.9)$ & $43.0$ \scriptsize$(\pm\,26.5)$ & $55.1$ \scriptsize$(\pm\,0.8)$ \\
PredRNet & $31.0$ \scriptsize$(\pm\,1.6)$ & $28.0$ \scriptsize$(\pm\,0.7)$ & $27.9$ \scriptsize$(\pm\,0.5)$ & $62.3$ \scriptsize$(\pm\,2.2)$ & $56.9$ \scriptsize$(\pm\,1.4)$ & $48.5$ \scriptsize$(\pm\,0.9)$ \\
RelBase & $36.6$ \scriptsize$(\pm\,0.8)$ & $29.7$ \scriptsize$(\pm\,0.6)$ & $31.1$ \scriptsize$(\pm\,1.0)$ & $73.0$ \scriptsize$(\pm\,1.8)$ & $66.2$ \scriptsize$(\pm\,1.0)$ & $\underline{65.7}$ \scriptsize$(\pm\,4.6)$ \\
SCL & $\underline{40.8}$ \scriptsize$(\pm\,3.2)$ & $\underline{32.0}$ \scriptsize$(\pm\,2.3)$ & $\textbf{33.5}$ \scriptsize$(\pm\,0.7)$ & $\underline{75.6}$ \scriptsize$(\pm\,10.1)$ & $60.0$ \scriptsize$(\pm\,4.1)$ & $63.9$ \scriptsize$(\pm\,4.3)$ \\
SRAN & $22.7$ \scriptsize$(\pm\,1.1)$ & $20.9$ \scriptsize$(\pm\,0.9)$ & $23.3$ \scriptsize$(\pm\,0.3)$ & $42.1$ \scriptsize$(\pm\,2.3)$ & $39.9$ \scriptsize$(\pm\,2.7)$ & $34.6$ \scriptsize$(\pm\,3.6)$ \\
STSN & $27.3$ \scriptsize$(\pm\,4.6)$ & $21.9$ \scriptsize$(\pm\,4.6)$ & $12.3$ \scriptsize$(\pm\,0.1)$ & $39.9$ \scriptsize$(\pm\,14.7)$ & $25.7$ \scriptsize$(\pm\,10.6)$ & $20.7$ \scriptsize$(\pm\,7.7)$ \\
WReN & $13.5$ \scriptsize$(\pm\,0.1)$ & $13.8$ \scriptsize$(\pm\,0.7)$ & $14.1$ \scriptsize$(\pm\,0.2)$ & $18.0$ \scriptsize$(\pm\,0.4)$ & $17.1$ \scriptsize$(\pm\,0.2)$ & $17.7$ \scriptsize$(\pm\,0.6)$ \\
\midrule
PoNG (ours) & $\textbf{44.7}$ \scriptsize$(\pm\,2.1)$ & $\textbf{34.3}$ \scriptsize$(\pm\,0.8)$ & $\underline{32.1}$ \scriptsize$(\pm\,2.1)$ & $\textbf{81.4}$ \scriptsize$(\pm\,3.1)$ & $\textbf{70.0}$ \scriptsize$(\pm\,4.1)$ & $\textbf{81.3}$ \scriptsize$(\pm\,1.6)$ \\
\bottomrule
\end{tabular}
\caption{\textbf{A-I-RAVEN extended regimes.} P, A, and D3 denote Progression, Arithmetic, and Distribute Three, resp.}
\label{tab:attributeless-extended}
\end{table*}
% \addtolength{\tabcolsep}{1pt}

%% file: tables/pgm.tex
% \addtolength{\tabcolsep}{-1pt}
\begin{table*}[t]
\centering
\small
\begin{tabular}{lccccccccc}
\toprule
Model & Neutral & Interpolation & HO-AP & HO-TP & HO-Triples & HO-LT & HO-SC & Extrapolation & Average \\
\midrule
SCL & $87.1$ & $56.0$ & $79.6$ & $76.6$ & $23.0$ & $14.1$ & $12.6$ & $19.8$ & $46.1$ \\
MRNet & $93.4$ & $68.1$ & $38.4$ & $55.3$ & $25.9$ & $\textbf{30.1}$ & $\textbf{16.9}$ & $19.2$ & $43.4$ \\
ARII & $88.0$ & $57.8$ & $50.0$ & $64.1$ & $32.1$ & $16.0$ & $12.7$ & $\underline{29.0}$ & $43.7$ \\
PredRNet & $97.4$ & $70.5$ & $63.4$ & $67.8$ & $23.4$ & $27.3$ & $13.1$ & $19.7$ & $47.8$ \\
DRNet & $\textbf{99.1}$ & $\underline{83.8}$ & $\textbf{93.7}$ & $78.1$ & $\textbf{48.8}$ & $\underline{27.9}$ & $13.1$ & $22.2$ & $\textbf{58.3}$ \\
Slot-Abstractor & $91.5$ & $\textbf{91.6}$ & $63.3$ & $\underline{78.3}$ & $20.4$ & $16.7$ & $\underline{14.3}$ & $\textbf{39.3}$ & $51.9$ \\
PoNG (ours) & $\underline{98.1}$ & $75.2$ & $\underline{92.1}$ & $\textbf{97.7}$ & $\underline{46.1}$ & $16.9$ & $12.6$ & $19.9$ & $\underline{57.3}$ \\
\bottomrule
\end{tabular}
\caption{\textbf{PGM}. Test accuracy of PoNG in all regimes of the PGM dataset. The Held-out Attribute Pairs regime is denoted as HO-AP, Held-out Triple Pairs as HO-TP, Held-out Triples as HO-Triples, Held-out Attribute line-type as HO-LT, and Held-out Attribute shape-colour as HO-SC. For reference, we provide results of SCL~\protect\cite{wu2020scattering,malkinski2020multi}, MRNet~\protect\cite{benny2020scale}, ARII~\protect\cite{zhang2022learningrobust}, PredRNet~\protect\cite{yang2023neural}, DRNet~\protect\cite{zhao2024learning}, and Slot-Abstractor~\protect\cite{mondal2024slot}.}
\label{tab:pgm}
\end{table*}
% \addtolength{\tabcolsep}{1pt}

%% file: tables/vap.tex
% \addtolength{\tabcolsep}{-1pt}
\begin{table*}[t]
    \centering
    \small
    \begin{tabular}{llllllc}
        \toprule
         & ND Transfer & NTD LineType & NTD ShapeColor & NAV Interpolation & NAV Extrapolation & Average \\
        \midrule
        LBC & $0.87 \pm 0.005$ & $0.76 \pm 0.020$ & $0.78 \pm 0.004$ & $0.93 \pm 0.004$ & $0.62 \pm 0.020$ & $0.79$ \\
        NSM & $0.88$ & $\underline{0.79}$ & $0.78$ & $0.93$ & $\textbf{0.74}$ & $0.82$ \\
        PredRNet & $\underline{0.96} \pm 0.003$ & $\textbf{0.82} \pm 0.010$ & $\underline{0.80} \pm 0.010$ & $\underline{0.97} \pm 0.002$ & $\underline{0.72} \pm 0.060$ & $\textbf{0.85}$ \\
        PoNG (ours) & $\textbf{0.98} \pm 0.001$ & $0.78 \pm 0.006$ & $\textbf{0.81} \pm 0.006$ & $\textbf{0.98} \pm 0.000$ & $0.68 \pm 0.007$ & $\underline{0.84}$ \\
        \bottomrule
    \end{tabular}
    \caption{\textbf{Visual Analogy Problems~\protect\cite{hill2018learning}.} Results of LBC, NSM, and PredRNet come from~\protect\cite[Table~2d]{yang2023neural}. For PoNG, we present mean and std of test accuracy for three random seeds. ND denotes Novel Domain, NTD — Novel Target Domain, NAV — Novel Attribute Values.}
    \label{tab:vap}
\end{table*}
% \addtolength{\tabcolsep}{1pt}

%% file: tables/vasr.tex
% \addtolength{\tabcolsep}{-1pt}
\begin{table*}[t]
    \centering
    \small
    \begin{tabular}{lccccc}
        \toprule
        Distractors & Zero-Shot ViT & Zero-Shot Swin & Supervised Concat & PoNG (best-of-3) & PoNG (mean $\pm$ std) \\
        \midrule
        Random & $86.0$ & $86.0$ & $84.1$ & $\textbf{92.0}$ & $91.8 \pm 0.3$ \\
        Difficult & $50.3$ & $52.9$ & $54.9$ & $\textbf{70.5}$ & $69.5 \pm 1.1$ \\
        \bottomrule
    \end{tabular}
    \caption{\textbf{Visual Analogies of Situation Recognition (VASR)}~\protect\cite{bitton2023vasr}. Results of selected baselines come from~\protect\cite[Table~3]{bitton2023vasr}.
    For PoNG, we present mean with std and best-of-3 test accuracy for three random seeds.}
    \label{tab:vasr}
\end{table*}
% \addtolength{\tabcolsep}{1pt}

%% file: tables/ablations.tex
% \addtolength{\tabcolsep}{-2pt}
\begin{table*}[t]
\centering
\small
\begin{tabular}{lcccccc}
\toprule
 & I-RAVEN & I-RAVEN-Mesh & \texttt{A/Color} & \texttt{A/Position} & \texttt{A/Size} & \texttt{A/Type} \\
\midrule
w/o P1 and P2 & $92.8$ \scriptsize$(-\ \ 3.1)$ & $74.4$ \scriptsize$(-14.9)$ & $73.3$ \scriptsize$(-\ \ 7.0)$ & $76.4$ \scriptsize$(-\ \ 2.9)$ & $58.4$ \scriptsize$(-15.2)$ & $49.5$ \scriptsize$(-\ \ 9.8)$ \\
w/o P3 and P4 & $95.6$ \scriptsize$(-\ \ 0.3)$ & $88.0$ \scriptsize$(-\ \ 1.3)$ & $78.9$ \scriptsize$(-\ \ 1.4)$ & $78.6$ \scriptsize$(-\ \ 0.7)$ & $73.9$ \scriptsize$(+\ \ 0.4)$ & $53.9$ \scriptsize$(-\ \ 5.5)$ \\
w/o TCN & $96.0$ \scriptsize$(+\ \ 0.1)$ & $90.8$ \scriptsize$(+\ \ 1.4)$ & $75.4$ \scriptsize$(-\ \ 4.9)$ & $80.3$ \scriptsize$(+\ \ 1.0)$ & $66.6$ \scriptsize$(-\ \ 6.9)$ & $57.5$ \scriptsize$(-\ \ 1.9)$ \\
$\beta=0$ & $94.2$ \scriptsize$(-\ \ 1.7)$ & $91.4$ \scriptsize$(+\ \ 2.1)$ & $79.0$ \scriptsize$(-\ \ 1.3)$ & $77.5$ \scriptsize$(-\ \ 1.8)$ & $70.3$ \scriptsize$(-\ \ 3.2)$ & $53.3$ \scriptsize$(-\ \ 6.1)$ \\
$\gamma=0$ & $95.7$ \scriptsize$(-\ \ 0.1)$ & $88.8$ \scriptsize$(-\ \ 0.5)$ & $74.2$ \scriptsize$(-\ \ 6.1)$ & $79.6$ \scriptsize$(+\ \ 0.3)$ & $73.0$ \scriptsize$(-\ \ 0.5)$ & $56.9$ \scriptsize$(-\ \ 2.5)$ \\
$\beta=0 \land \gamma=0$ & $79.7$ \scriptsize$(-16.2)$ & $32.7$ \scriptsize$(-56.7)$ & $72.1$ \scriptsize$(-\ \ 8.2)$ & $75.1$ \scriptsize$(-\ \ 4.2)$ & $64.9$ \scriptsize$(-\ \ 8.6)$ & $49.0$ \scriptsize$(-10.3)$ \\
union & $81.4$ \scriptsize$(-14.5)$ & $32.5$ \scriptsize$(-56.8)$ & $76.2$ \scriptsize$(-\ \ 4.1)$ & $74.1$ \scriptsize$(-\ \ 5.2)$ & $66.9$ \scriptsize$(-\ \ 6.6)$ & $46.0$ \scriptsize$(-13.4)$ \\
\bottomrule
\end{tabular}
\caption{\textbf{PoNG ablations}. Test accuracy averaged across 3 random seeds and a difference to the default model setup (cf. Table~\ref{tab:attributeless}). Union denotes application of all ablations except for the first one.}
\label{tab:ablations}
\end{table*}
% \addtolength{\tabcolsep}{2pt}

%% file: sections/conclusion.tex
\section{Conclusion}
\label{sec:conclusion}

Generalization to novel problem types is an active and open area of DL research.
In this work, we introduced PoNG, a novel AVR model that leverages group convolution, parallel design, weight sharing, and normalization.
To evaluate its effectiveness and versatility, we conducted experiments on four RPM benchmarks and two visual analogy datasets comprising both synthetic and real-world images.
PoNG demonstrates strong performance across all considered problems, often surpassing the state-of-the-art reference methods.

\paragraph{Future directions.}
We believe that the proposed pathways block, a key component of PoNG, is a generic module also applicable to other tasks that require reasoning over a set of objects (vector embeddings).
Nevertheless, the presented experimental evaluation of PoNG is focused on variable RPM benchmarks, including I-RAVEN, I-RAVEN-Mesh, A-I-RAVEN, and PGM, and two visual analogy datasets, i.e. VAP and VASR.
Assessing the model's performance on problems outside the AVR domain constitutes an interesting continuation of this work.

%% file: sections/acknowledgment.tex
\section*{Acknowledgments}
This research was carried out with the support of the Laboratory of Bioinformatics and Computational Genomics and the High Performance Computing Center of the Faculty of Mathematics and Information Science Warsaw University of Technology.
Mikołaj Małkiński was funded by the Warsaw University of Technology within the Excellence Initiative: Research University (IDUB) programme.

%% file: sections/appendix.tex
\section{Model details}
\label{sec:model-details}
Tables~\ref{tab:pong-hparams-panel-encoder},~\ref{tab:pong-hparams-reasoner}, and~\ref{tab:pong-hparams-prediction-heads} list all PoNG hyperparameters.
Table~\ref{tab:size} compares the parameter counts of the evaluated models.

\section{Qualitative analysis}
\label{sec:qualitative-analysis}

TCN operates over task-relevant groups to preserve the relations between representations within a group, while discarding absolute magnitude, which we hypothesize helps extrapolation to novel input domains.
In the first Pathways block, TCN is applied to row embeddings (P3) and group-pair embeddings (P4), supporting row-wise rule recognition, particularly in out-of-distribution settings.
Ablations in Section~\ref{sec:experiments} confirm this -- removing TCN degrades performance in settings with larger attribute domains (e.g.\ \texttt{A/Color} and \texttt{A/Size}; see Table~\ref{tab:ablations}, "w/o TCN").

To better understand the role of TCN in PoNG, we performed error analysis referring to the impact of TCN on PoNG predictions on these two datasets.
Figs.~\ref{fig:tcn-color} and~\ref{fig:tcn-size} illustrate failure cases where the ablation model (PoNG without TCN) selected incorrect answers differing only in the held-out attribute (Color or Size, resp.), indicating a failure to generalize attribute-specific rules.
Table~\ref{tab:metrics} further quantifies this degradation -- the ablation variant exhibits worse cross-entropy (CE), total variation distance (TVD), and Brier scores on test splits of both \texttt{A/Color} and \texttt{A/Size} datasets.
This analysis suggests that TCN plays an important role in supporting rule generalization to novel attributes. 

\input{tables/pong-hparams}
\input{figures/tcn-color}
\input{figures/tcn-size}
\input{tables/model-size}
\input{tables/metrics-tcn}
% \FloatBarrier
% \vfill
% \vfill

%% file: tables/pong-hparams.tex
\begin{table}
    \centering
    \small
    \begin{sc}
        \begin{tabular}{lc}
            \toprule
            Layer & Hyperparameters \\
            \midrule

            % Pathway 1
            Conv2D-ReLU-BN2D & $[1 \to 32, 7 \times 7, 2 \times 2, 3 \times 3]$ \\
            Conv2D-ReLU-BN2D & $[32 \to 32, 7 \times 7, 2 \times 2, 3 \times 3]$ \\
            % Pathway 2
            MaxPool & $[3 \times 3, 2 \times 2, 1 \times 1]$ \\
            MaxPool & $[3 \times 3, 2 \times 2, 1 \times 1]$ \\
            Conv2D & $[1 \to 32, 1 \times 1, 1 \times 1, 0 \times 0]$ \\
            Sum & \\
            \midrule
            % Pathway 1
            Conv2D-ReLU-BN2D & $[32 \to 32, 7 \times 7, 2 \times 2, 3 \times 3]$ \\
            Conv2D-ReLU-BN2D & $[32 \to 32, 7 \times 7, 2 \times 2, 3 \times 3]$ \\
            % Pathway 2
            MaxPool & $[3 \times 3, 2 \times 2, 1 \times 1]$ \\
            MaxPool & $[3 \times 3, 2 \times 2, 1 \times 1]$ \\
            Conv2D & $[32 \to 32, 1 \times 1, 1 \times 1, 0 \times 0]$ \\
            Sum & \\
            \midrule

            % Flatten (height \& width) & $[5 \times 5 \to 25]$ \\
            Flatten & $[5 \times 5 \to 25]$ \\
            Linear-ReLU & $[25 \to 25]$ \\
            % Flatten (depth \& spatial) & $[32 \times 25 \to 800]$ \\
            Flatten & $[32 \times 25 \to 800]$ \\
            LN & \\
            Linear-ReLU & $[800 \to 1600]$ \\
            LN & \\
            Linear & $[1600 \to 800]$ \\
            Sum & \\
            Position embedding & $[25]$ \\
            % Position embedding & $[800 \to 825]$ \\

            \bottomrule
        \end{tabular}
    \end{sc}
    \caption{\textbf{PoNG hyperparameters: Panel encoder $\mathcal{E}$.}
    The parameters
    of convolution layers are denoted as $[$\#~input channels $\to$ \#~output channels, kernel size, stride, padding$]$;
    of pooling layers as $[$kernel size, stride, padding$]$;
    of linear layers as $[$\#~input neurons $\to$ \#~output neurons$]$;
    of flatten operators as $[$input dimensions $\to$ output dimensions$]$;
    of position embedding as $[$dimensionality of the position embedding vector$]$.}
    \label{tab:pong-hparams-panel-encoder}
\end{table}

\begin{table}
    \centering
    \small
    \begin{sc}
        \begin{tabular}{lc}
            \toprule
            Layer & Hyperparameters \\
            \midrule

            % Stack & $\to [B, 9, 825]$ \\
            Stack & \\

            % === 1st Pathways block ===

            % \midrule
            BN1D & \\
            % Pathway 1
            Conv1D & $[9 \to 32, 1, 1, 0, \text{bias} = 0]$ \\
            % Pathway 2
            Conv1D-ReLU-BN1D & $[9 \to 32, 7, 1, 3]$ \\
            % Pathway 3
            G-C-TCN-ReLU-BN1D & $[3 \to 32, 7, 1, 3, 3]$ \\
            % Pathway 4
            GP-C-TCN-ReLU-BN1D & $[6 \to 32, 7, 1, 3, 3]$ \\

            % === 2nd Pathways block ===

            \midrule
            % Bottleneck
            AvgPool1D & $[10, 8, 1]$ \\
            BN1D & \\
            % Pathway 1
            Conv1D & $[32 \to 32, 1, 1, 0, \text{bias} = 0]$ \\
            % Pathway 2
            Conv1D-ReLU-BN1D & $[32 \to 32, 7, 1, 3]$ \\
            Conv1D-ReLU-BN1D & $[32 \to 32, 7, 1, 3]$ \\
            % Pathway 3
            G-C-TCN-ReLU-BN1D & $[4 \to 32, 7, 1, 3, 8]$ \\
            G-C-TCN-ReLU-BN1D & $[4 \to 32, 7, 1, 3, 8]$ \\
            % Pathway 4
            GP-C-TCN-ReLU-BN1D & $[16 \to 32, 7, 1, 3, 4]$ \\
            GP-C-TCN-ReLU-BN1D & $[16 \to 32, 7, 1, 3, 4]$ \\

            % === 3rd Pathways block ===

            \midrule
            % Bottleneck
            AvgPool1D & $[6, 4, 1]$ \\
            BN1D & \\
            % Pathway 1
            Conv1D & $[32 \to 32, 1, 1, 0, \text{bias} = 0]$ \\
            % Pathway 2
            Conv1D-ReLU-BN1D & $[32 \to 32, 7, 1, 3]$ \\
            Conv1D-ReLU-BN1D & $[32 \to 32, 7, 1, 3]$ \\
            % Pathway 3
            G-C-TCN-ReLU-BN1D & $[4 \to 32, 7, 1, 3, 8]$ \\
            G-C-TCN-ReLU-BN1D & $[4 \to 32, 7, 1, 3, 8]$ \\
            % Pathway 4
            GP-C-TCN-ReLU-BN1D & $[16 \to 32, 7, 1, 3, 4]$ \\
            GP-C-TCN-ReLU-BN1D & $[16 \to 32, 7, 1, 3, 4]$ \\

            \midrule
            Adaptive AvgPool1D & $[25 \to 16]$ \\
            % Flatten (depth \& feature dim) & $[32 \times 16 \to 512]$ \\
            Flatten & $[32 \times 16 \to 512]$ \\
            Linear-ReLU-BN1D & $[512 \to 512]$ \\
            Linear & $[512 \to 128]$ \\

            \bottomrule
        \end{tabular}
    \end{sc}
    \caption{\textbf{PoNG hyperparameters: Reasoner $\mathcal{R}$.}
    The parameters
    of convolution layers are denoted as $[$\#~input channels $\to$ \#~output channels, kernel size, stride, padding$]$;
    of group and group-pair convolution layers as $[$\#~input channels $\to$ \#~output channels, kernel size, stride, padding, groups$]$;
    of pooling layers as $[$kernel size, stride, padding$]$;
    of linear layers as $[$\#~input neurons $\to$ \#~output neurons$]$;
    of flatten operators as $[$input dimensions $\to$ output dimensions$]$.}
    \label{tab:pong-hparams-reasoner}
\end{table}

\begin{table}
    \centering
    \small
    \begin{sc}
        \begin{tabular}{lc}
            \toprule
            Layer & Hyperparameters \\

            \midrule
            \multicolumn{2}{l}{Target head $\mathcal{P}^y$} \\
            \midrule
            Linear-ReLU-Linear & $[128 \to 128]$ \\
            Linear & $[128 \to 1]$ \\

            \midrule
            \multicolumn{2}{l}{Aggregate rule head $\mathcal{P}_1^r$} \\
            \midrule
            Sum & \\
            Linear-ReLU & $[128 \to 128]$ \\
            Linear & $[128 \to d_r]$ \\

            \midrule
            \multicolumn{2}{l}{Target-conditioned rule head $\mathcal{P}_2^r$} \\
            \midrule
            Weighted sum & \\
            Linear & $[128 \to d_r]$ \\

            \bottomrule
        \end{tabular}
    \end{sc}
    \caption{\textbf{PoNG hyperparameters: Prediction heads.}
    The parameters
    of linear layers are denoted as $[$\#~input neurons $\to$ \#~output neurons$]$.}
    \label{tab:pong-hparams-prediction-heads}
\end{table}

%% file: figures/tcn-color.tex
\begin{figure*}[!h]
    \centering
    \begin{subfigure}{0.22\textwidth}
        \centering
        \includegraphics[width=\linewidth]{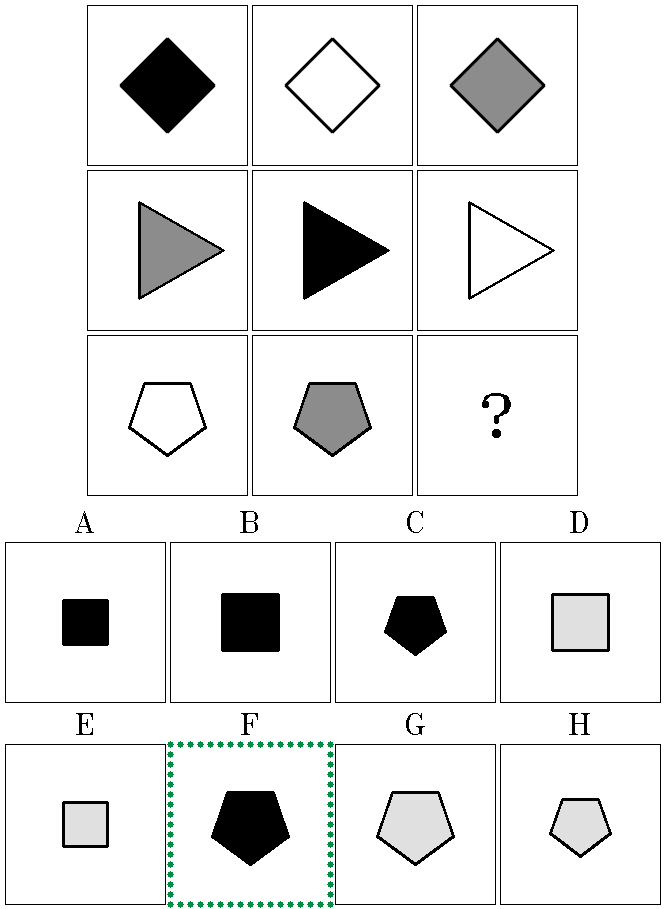}
        \includegraphics[width=\linewidth]{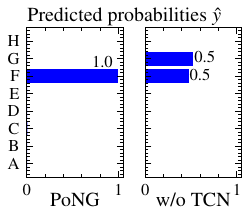}
        \caption{\texttt{Center}}
    \end{subfigure}
    \hfil
    \begin{subfigure}{0.22\textwidth}
        \centering
        \includegraphics[width=\linewidth]{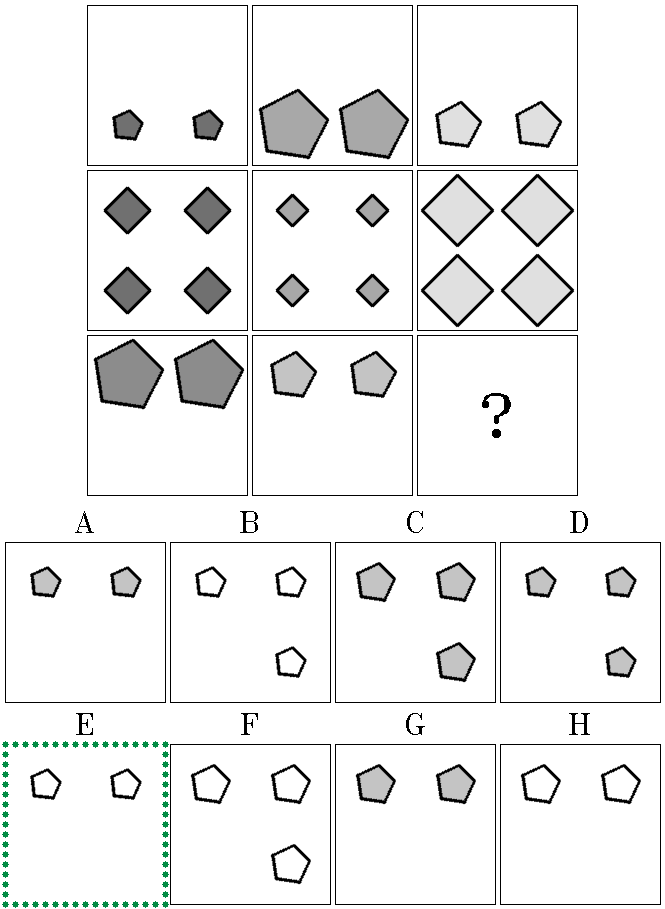}
        \includegraphics[width=\linewidth]{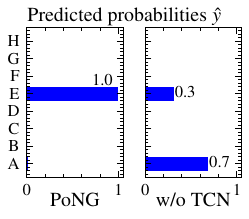}
        \caption{\texttt{2x2Grid}}
    \end{subfigure}
    \hfil
    \begin{subfigure}{0.22\textwidth}
        \centering
        \includegraphics[width=\linewidth]{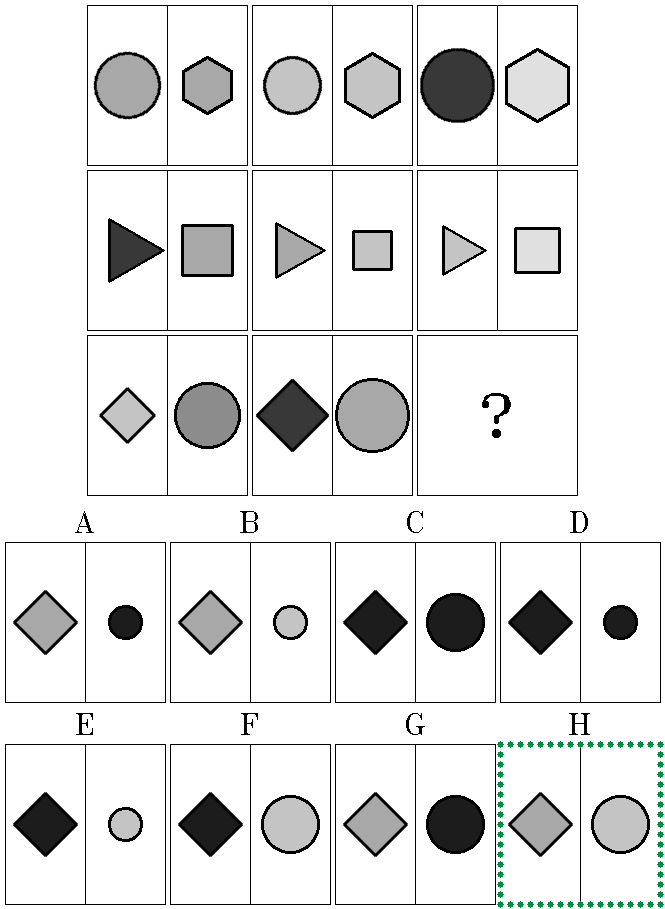}
        \includegraphics[width=\linewidth]{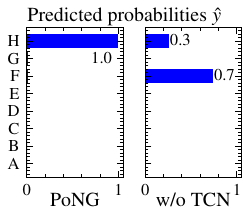}
        \caption{\texttt{Left-Right}}
    \end{subfigure}
    \hfil
    \begin{subfigure}{0.22\textwidth}
        \centering
        \includegraphics[width=\linewidth]{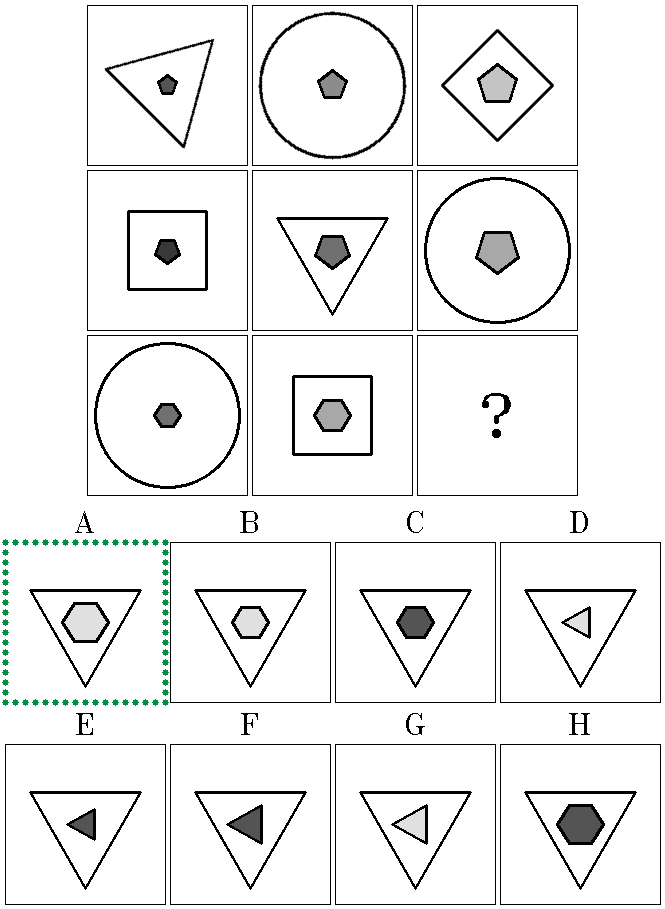}
        \includegraphics[width=\linewidth]{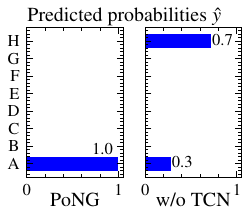}
        \caption{\texttt{Out-InCenter}}
    \end{subfigure}
    \caption{
    Selected examples of \texttt{A/Color} where PoNG succeeded and its ablation variant without TCN failed.
    In each case, the ablation model selected an answer differing in the held-out Color attribute.
    Missed rules:
    (a) \textit{Same set of $3$ colors per row.}
    (b) \textit{Decreasing color progression from left to right.}
    (c) \textit{Consistent set of $3$ colors in the left image part across rows.}
    (d) \textit{Decreasing color progression in the inner figure from left to right.}
    }
    \label{fig:tcn-color}
\end{figure*}

%% file: figures/tcn-size.tex
\begin{figure*}[!h]
    \centering
    \begin{subfigure}{0.22\textwidth}
        \centering
        \includegraphics[width=\linewidth]{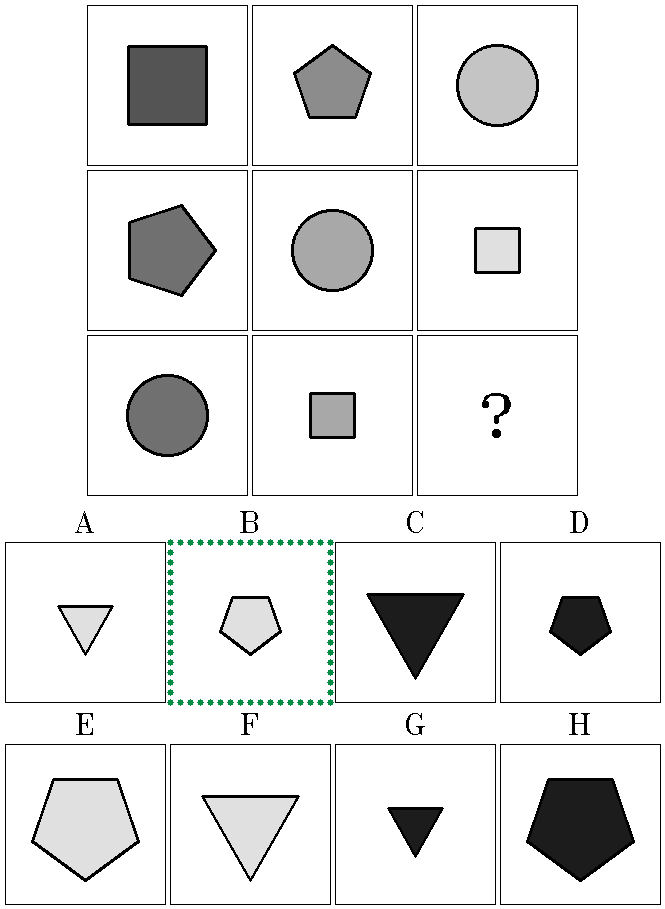}
        \includegraphics[width=\linewidth]{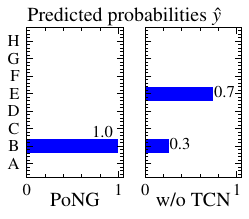}
        \caption{\texttt{Center}}
    \end{subfigure}
    \hfil
    \begin{subfigure}{0.22\textwidth}
        \centering
        \includegraphics[width=\linewidth]{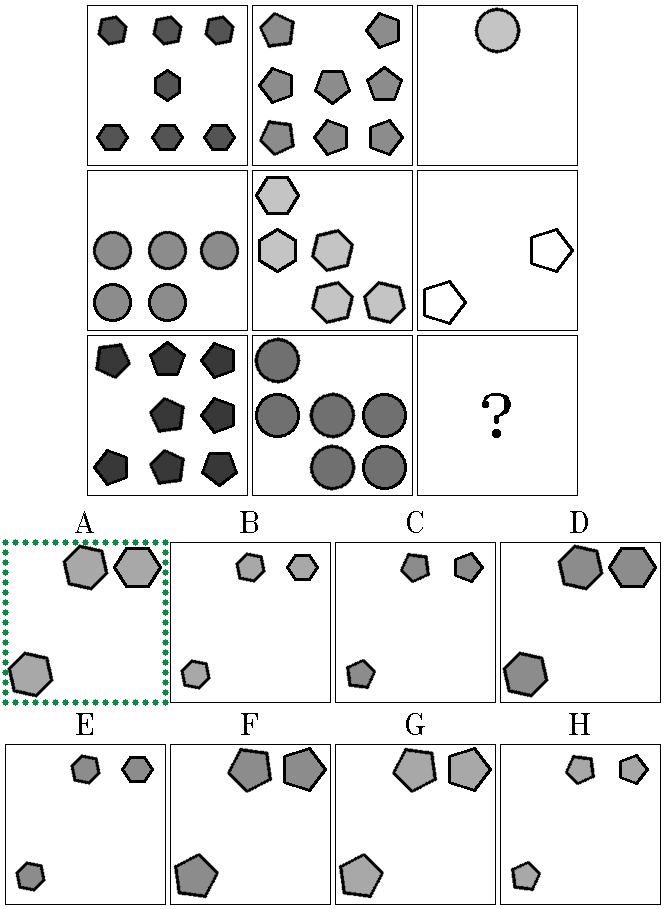}
        \includegraphics[width=\linewidth]{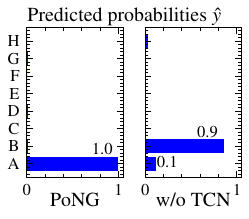}
        \caption{\texttt{3x3Grid}}
    \end{subfigure}
    \hfil
    \begin{subfigure}{0.22\textwidth}
        \centering
        \includegraphics[width=\linewidth]{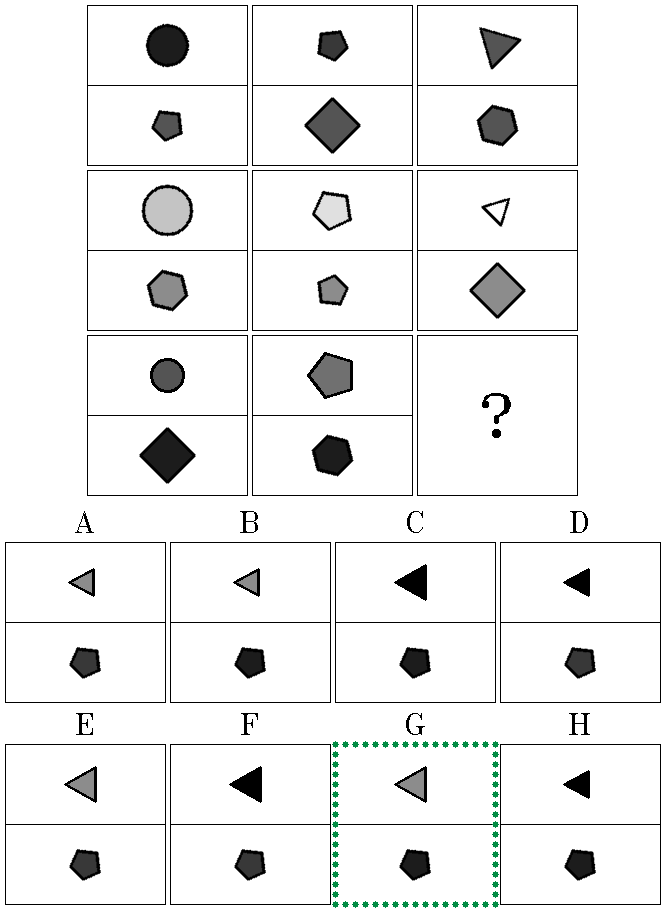}
        \includegraphics[width=\linewidth]{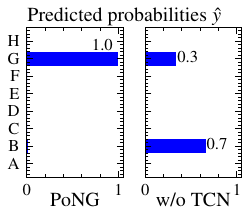}
        \caption{\texttt{Up-Down}}
    \end{subfigure}
    \hfil
    \begin{subfigure}{0.22\textwidth}
        \centering
        \includegraphics[width=\linewidth]{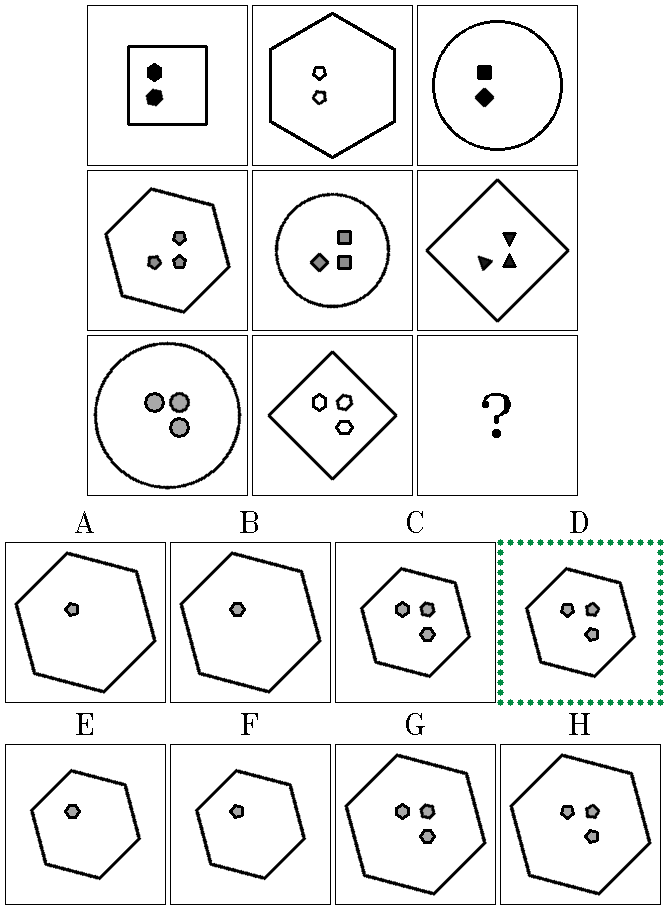}
        \includegraphics[width=\linewidth]{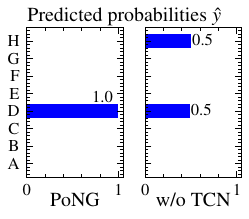}
        \caption{\texttt{Out-InGrid}}
    \end{subfigure}
    \caption{
    Selected examples of \texttt{A/Size} where PoNG succeeded and its ablation variant without TCN failed.
    In each case, the ablation model selected an answer differing in the held-out Size attribute.
    Missed rules:
    (a) \textit{Decreasing size from left to right.}
    (b) \textit{Constant size across rows.}
    (c) \textit{Same set of $3$ sizes in the lower image part across rows.}
    (d) \textit{Consistent set of $3$ sizes for the outer figure.}
    }
    \label{fig:tcn-size}
\end{figure*}

%% file: tables/model-size.tex
\begin{table}[H]
    \centering    
     \begin{sc}
        \small
        \begin{tabular}{lr|lr}
            \toprule
            Model & \# Params & Model & \# Params\\
            \midrule
            SCL & $0.6$M & CPCNet & $3.9$M \\
            STSN & $1.0$M & MRNet & $5.2$M \\
            RelBase & $1.3$M & CNN-LSTM & $6.7$M \\
            CoPINet & $1.7$M & WReN & $13.2$M \\
            PredRNet & $2.0$M & DRNet & $24.7$M \\
            PoNG & $3.1$M & SRAN & $45.7$M \\
            \bottomrule
        \end{tabular}
        \caption{The number of model parameters in millions (M).}
        \label{tab:size}
    \end{sc}   
\end{table}

%% file: tables/metrics-tcn.tex
\addtolength{\tabcolsep}{-1.5pt}
\begin{table}[H]
    \centering
    \small
    \begin{tabular}{lcccccc}
        \toprule
         & \multicolumn{3}{c}{\texttt{A/Color}} & \multicolumn{3}{c}{\texttt{A/Size}} \\
         \cmidrule(lr){2-4} \cmidrule(lr){5-7}
         & CE $\downarrow$ & TVD $\downarrow$ & Brier $\downarrow$ & CE $\downarrow$ & TVD $\downarrow$ & Brier $\downarrow$ \\
        \midrule
        PoNG & $1.503$ & $3\,370$ & $4\,812$ & $1.536$ & $3\,836$ & $5\,754$ \\
        w/o TCN & $1.566$ & $4\,212$ & $7\,066$ & $1.550$ & $4\,037$ & $6\,137$ \\
        \bottomrule
    \end{tabular}
    \caption{Cross-entropy (CE), total variation distance (TVD), and Brier score for PoNG and its ablation variant without TCN (w/o TCN) on the full \texttt{A/Color} and \texttt{A/Size} test sets.}
    \label{tab:metrics}
\end{table}
\addtolength{\tabcolsep}{-1.5pt}